%% file: EURO2024_arXiv_version0.tex
\documentclass[12pt]{article}

\usepackage{graphicx}
\usepackage{multicol}
\usepackage{amsmath}
\usepackage{amssymb}

\usepackage{geometry}
\usepackage{color}
\usepackage{bbold}
\usepackage{lscape}
\usepackage{setspace}
\usepackage{array}
\usepackage{multirow}
\usepackage{booktabs}
\usepackage{geometry}
\usepackage[font=small,labelfont=bf]{caption}
\usepackage{subfig}
\usepackage{fancybox}
\usepackage{enumitem}
\usepackage{tabularx}
\usepackage{float}
\usepackage{ifthen}
\usepackage{amsfonts}
\usepackage{amsthm}
\usepackage{eurosym}
\usepackage{setspace}
\usepackage{array}
\usepackage{multirow}
\usepackage{hyperref}
\usepackage{footmisc}

\usepackage{dsfont}
\usepackage[table]{xcolor}
\usepackage{lscape}
\usepackage{savefnmark}
\usepackage{pdfpages}

\newcommand{\wb}{\textbf{w}}

\newcommand{\xb}{\textbf{x}}

\newcommand{\blanco}[1]{}

\makeatletter
\def\maxwidth{ %
  \ifdim\Gin@nat@width>\linewidth
    \linewidth
  \else
    \Gin@nat@width
  \fi
}
\makeatother

\definecolor{fgcolor}{rgb}{0.345, 0.345, 0.345}

\usepackage{algorithm}
\usepackage[noend]{algpseudocode}

\makeatletter
\def\BState{\State\hskip-\ALG@thistlm}
\makeatother

\usepackage{framed}
\makeatletter
 {\par\unskip\endMakeFramed%
 \at@end@of@kframe}
\makeatother

\definecolor{shadecolor}{rgb}{.97, .97, .97}
\definecolor{messagecolor}{rgb}{0, 0, 0}
\definecolor{warningcolor}{rgb}{1, 0, 1}
\definecolor{errorcolor}{rgb}{1, 0, 0}
\usepackage{alltt}

\usepackage{mathptmx}

\usepackage{graphics}

\usepackage{natbib} 

\usepackage{tikz}
\usetikzlibrary{trees}
\usepackage{colortbl}
\usepackage{color}
\usepackage{booktabs}
\usepackage{hyperref}
\usepackage{amsmath}
\usepackage{amssymb}
\usepackage{float}
\usepackage{subfig}

\definecolor{lightgray}{rgb}{0.75, 0.75, 0.75}

\def\lc{\left\lfloor}   
\def\rc{\right\rfloor}

\begin{document}
	
\title{Modeling and Prediction of the UEFA EURO 2024 via Combined Statistical Learning Approaches}

\author{A. Groll
\thanks{Department of Statistics, TU Dortmund University, Vogelpothsweg 87, 44227 Dortmund, Germany, \emph{groll@statistik.tu-dortmund.de}}
\and L. M. Hvattum
\thanks{Molde University College, Molde, Norway, \emph{Lars.M.Hvattum@himolde.no}}
\and C. Ley
\thanks{Faculty of Sciences, Department of Applied Mathematics, Computer Science and Statistics, Ghent University, Krijgslaan 281, 9000 Gent,  Belgium, \emph{Christophe.Ley@UGent.be}}
\and J. Sternemann
\thanks{Department of Computer Science, TU Dortmund University, Otto-Hahn-Str. 14, 44227 Dortmund, Germany, \emph{jonas.sternemann@tu-dortmund.de}}
\and G. Schauberger
\thanks{Chair of Epidemiology, Department of Sport and Health Sciences, Technical University of Munich, \emph{g.schauberger@tum.de}}
%\and H. Van Eetvelde
%\thanks{Faculty of Sciences, Department of Applied Mathematics, Computer Science and Statistics, Ghent University, Krijgslaan 281, 9000 Gent,  Belgium, \emph{hans.vaneetvelde@ugent.be}}
\and A. Zeileis
\thanks{Department of Statistics, Universit\"at Innsbruck, Austria, \emph{ Achim.Zeileis@R-project.org}}
}

\maketitle

\thispagestyle{empty}

\setlength{\parindent}{0pt}

\setlength{\columnsep}{15pt}

\textbf{Abstract}
{In this work, three fundamentally different machine learning models are combined to create a new, joint model for forecasting the UEFA EURO 2024. Therefore, a generalized linear model, a random forest model, and a extreme gradient boosting model are used to predict the number of goals a team scores in a match. The three models are trained on the match results of the UEFA EUROs 2004-2020, with additional covariates characterizing the teams for each tournament as well as three enhanced variables derived from different ranking methods for football teams. The first enhanced variable is based on historic match data from national teams, the second is based on the bookmakers' tournament winning odds of all participating teams, and the third is based on historic match data of individual players both for club and international matches, resulting in player ratings. Then, based on current covariate information of the participating teams, the final trained model is used to predict the UEFA EURO 2024. For this purpose, the tournament is simulated 100.000 times, based on the estimated expected number of goals for all possible matches, from which probabilities across the different tournament stages are derived. Our combined model identifies France as the clear favourite with a winning probability of 19.2\%, followed by England (16.7\%) and host Germany (13.7\%).
}
\bigskip

\bigskip

%\textbf{Disclaimer}:
%This manuscript strongly builds on a previous publication on modeling and predicting the UEFA EURO 2020, and is therefore in some aspects similar to \cite{groll:21}.

\section{Introduction}
Football is one of the most popular sports in the world, with particularly high interest of fans for tournaments like the FIFA World Cup or the UEFA European Championship (EURO). Hence, in advance of every major football tournament, discussions on which team is the favorite are very common among football enthusiasts, who are often biased towards their country. For that reason, having a more neutral and scientific view regarding this question has become more popular in recent years.
With the development of both statistical and machine learning models, some of them were used to predict international football tournaments. 

One basic approach is based on Poisson regression models, where it is assumed that the goals of two competing teams~$i,j\in\{1,\ldots,n\}$ are Poisson distributed, e.g.\ $X_{ij}\sim Po(\lambda_{ij})$ and $Y_{ij}\sim Po(\mu_{ij})$. Then the models predict a value for the intensity parameter of a Poisson distribution for each of the two competing teams, based on covariates characterizing the teams. The first to use these models where \citet{Dyte:2000}. They used the FIFA ranking together with the match venue to model the parameter of the Poisson distribution to simulate the FIFA World Cup 1998, assuming that the number of goals for the two competing teams is (conditionally) independent. Instead of using only one team strength variable \cite{GroAbe:2013} and \cite{GroSchTut:2015} introduced several additional covariates that possibly influence the teams' success, for UEFA EURO and FIFA World Cup data, respectively. Moreover, they used LASSO for regularization to avoid overfitting and to select only relevant variables for the prediction.

Another approach to model football tournaments is rather new, with the expected number of goals being modeled using random forests. These are a parallel ensemble of single regression trees and were firstly introduced by \cite{Breiman:2001}. The potential of using this model type for football tournament forecasting was explored by \cite{SchauGroll2018}, with additional comparison to classical Poisson regression methods, on data from the FIFA World Cups 2002-2014. It turned out that the random forests were beneficial and showed superior predictive performance. Along these lines, \cite{GroEtAl:WM2018b} and \cite{GroEtAl:WM2019} additionally added team abilities derived from ranking methods as covariates to the random forest, which further improved the predictive performance for both the men's FIFA World Cup 2018 and the women's FIFA World Cup 2019, respectively. Also \cite{groll:21} used a random forest to predict the UEFA EURO 2020, where the promising performance of the random forests was again displayed.

Another different modeling approach is {\it extreme gradient boosting} (XGBoost), introduced by \cite{chen2016xgboost}. The XGBoost method, in contrast to the random forests, is a sequential ensemble technique, which additively combines many weak learners resulting in a strong learner. This method was firstly investigated in \cite{groll:21} for predicting the UEFA EURO 2020. So at the end, we have three very different modeling approaches, which use covariate data as well as highly informative team abilities, obtained through separate statistical models. These methods can be used alone to create rankings for the teams, as well as to create new covariate data.

The first ranking method results in a team ability, which is based on historic match data, and is computed using a Poisson model (see \citealp{LeyWieEet2018}). Therefore, the historic matches are weighted based on several factors, e.g.\ the importance and the recency of a match.
The second is based on a approach introduced by \cite{Leit:2010a}, which uses the bookmakers' winning odds for each team to calculate winning probabilities. These are then used to estimate team abilities via ``inverse" tournament simulation such that the simulated winning probabilities are as close as possible to the ones obtained from the bookmakers' odds.
Finally, the last ranking method calculates average plus-minus player ratings, as proposed by \cite{Hv19} and further investigated in \cite{HvGe21}. These ratings are calculated for each player based on his team's performance on club and international level. Therefore, the historic match data includes further information about the current players on the pitch, goals scored, and red cards.

These three team abilities from the different ranking methods serve as additional covariates for our models, similar to the methodology in \cite{groll:21}. Additionally, we introduce a new model approach that combines the three models to generate a new prediction. This model should take advantage of the different modeling approaches, yielding better predictive performance.

The subsequent sections of the manuscript are arranged as follows. In Section~\ref{sec:data}, we describe the dataset used for modeling the UEFA EURO. This can be divided into two types, namely the preceding UEFA EUROs match results linked with covariate information and the enhanced variables. The latter are based on separate datasets and statistical models. One variable is based on an extensive set of historic match results from all national teams, the second is based on various bookmaker odds and the last is also constructed from historic match data (also on club level) with further information about the match including line-ups, goals and other in-game events. After that, in Section~\ref{sec:methods} we explain the ideas of different models, namely a LASSO-regularized Poisson regression model, random forests, extreme gradient boosting and our new combined model. Next, in Section~\ref{sec:combine}, we evaluate the predictive performance of the combined model against the individual models. Based on these results, in Section~\ref{sec:prediction} the best-performing model is fitted to the training data (UEFA EUROs 2004-2020), and shortly investigated regarding the variable importance of the model. Finally, the resulting estimates from the model are used to simulate the UEFA EURO 2024 100,000 times to calculate winning probabilities, as well as probabilities for reaching the single knockout stages for each team. Finally, Section~\ref{sec:conclusion} concludes.

%%%%%%%%%%%%%%%%%%%%%%%%%%%%
\section{Data}\label{sec:data}

In this section, we describe the data that is used to model and predict the UEFA EURO 2024, which is very similar to the data used in \cite{groll:21}. The data can be divided into three types, namely the (classical) covariate data containing variables that characterize the teams for the corresponding UEFA EURO. This data is then connected to the second type of data, including real match results of the corresponding UEFA EURO tournaments. The third type is represented by so-called enhanced variables, which themselves are based on other data, in order to estimate team strength variables. The first enhanced variable is an ability estimate for each team, which is based on weighted historic match results from the national teams \citep{LeyWieEet2018}. The second enhanced variable is based on the odds from various bookmakers for each team to win the tournament, from which another national team strength is obtained \citep{Leit:2010a}. Finally, the third enhanced variable is an average plus-minus player rating for every team, which is also based on historic match data, both on international and club level. The matches are characterized with further information, e.g.\ the players on the pitch, red cards, and goals scored \citep{PaHv20}.

\begin{figure}[!ht]
	\centering
	\includegraphics[width=1.1\textwidth]{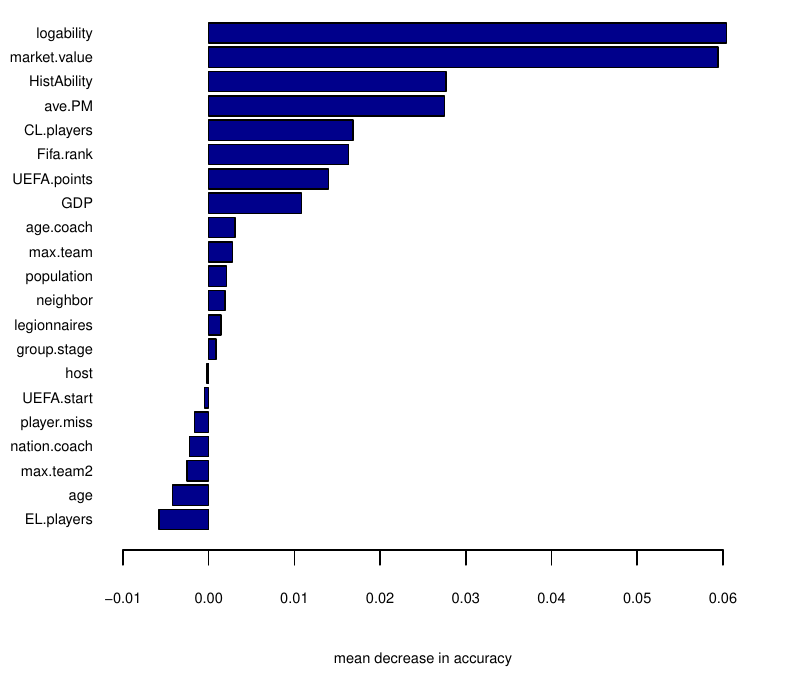}
	\caption{Bar plot showing the variable importance of a Cforest model (with \texttt{mtry} $=4$) trained on UEFA EURO 2004-2020 data.}
	\label{var_imp}
\end{figure}

In the previous predictions for the UEFA EURO \citep{GrollEtAl2018,groll:21} a set of around 20 variables was used for the models to predict the UEFA EURO. However, the resulting variable importance for the prediction of the UEFA EURO 2020 \citep{groll:21} already indicated that some variables are not very useful for the prediction. Besides the fact that the models already try to focus on the more important variables and partly are able to perform variable selection, we now exclude the irrelevant variables from the very beginning and only focus on the most important variables regarding a random forest model (in particular, a so-called Cforest model, which will be explained in more detail later on in Section~\ref{subsec:forest}) . The resulting variable importances of the Cforest model fitted to the UEFA EURO 2004-2020 data can be found in Figure~\ref{var_imp}.

It turns out that the logability and  market value are the most important ones, followed by the historic abilities and average plus-minus rating. So all three enhanced variables are among the top four most important variables. After the eight most important features there is an clear cut-off in the bar plot. For this reason, we choose only the top eight variables from Figure~\ref{var_imp} to model and predict the UEFA EURO 2024. This appears also to be more logical, as for some variables like the nationality of the coach it seems very likely that they have no relevant impact on the teams' performance.

As can be seen in Table~\ref{top8_comparison}, for almost all modeling approaches using only the top eight variables leads to better results and measures with respect to predictive performance. Only the Cforest model with a fixed \texttt{mtry} using the rule \texttt{mtry}$=\lc\sqrt{p}\rc$ (with $p$ being the number of covariates) has clearly better results using more variables. But the \textit{Cforest Tuned} model, which is used in this work, is better than the \textit{Cforest4} with \texttt{mtry}$= 4$. What also stands out is that the LASSO model profits the most from the reduction of the number of variables. Altogether, using only the top eight variables leads to better results. For more detailed information about the measures, see Section~\ref{sec:combine}, and for additional details about the models see Section~\ref{sec:methods}.
% latex table generated in R 4.3.3 by xtable 1.8-4 package
% Wed Jun 26 18:36:13 2024
\begin{table}[!ht]
	\centering
	\caption{
		\label{top8_comparison} 
		Model performances comparing the usage of the top eight features or all 21 features from \cite{groll:21}. For each pair of modeling approaches the better one is highlighted by bold font. \textit{Cforest4/Cforest2} means using a Cforest model with \texttt{mtry} = 4 or 2, while in \textit{Cforest Tuned} the parameter \texttt{mtry} is tuned.
	}
	\begin{tabular}{lccccc}
		\specialrule{.1em}{.05em}{.05em}
		& $MAE_{goals}$ & $MAE_{goaldiff}$ & ML & CR & RPS \\ 
		\specialrule{.1em}{.05em}{.05em}
		LASSO All & 0.8602 & 1.1857 & 0.3865 & 0.4821 & 0.2058 \\ 
		LASSO Top8 & \textbf{0.8550} & \textbf{1.1796} & \textbf{0.3983} & \textbf{0.4872} & \textbf{0.2028} \\ 
		\specialrule{.1em}{.05em}{.05em}
		Cforest4 All & \textbf{0.8671} & \textbf{1.1697} & 0.3959 & \textbf{0.4872} & \textbf{0.2030} \\ 
		Cforest2 Top8 & 0.8752 & 1.1794 & \textbf{0.3990} & 0.4821 & 0.2038 \\ 
		\specialrule{.1em}{.05em}{.05em}
		Cforest Tuned All & \textbf{0.8660} & 1.1735 & 0.3912 & \textbf{0.4923} & 0.2035 \\ 
		Cforest Tuned Top8 & 0.8686 & \textbf{1.1669} & \textbf{0.3996} & 0.4872 & \textbf{0.2016} \\ 
		\specialrule{.1em}{.05em}{.05em}
		XGBoost All & 0.8799 & 1.1970 & 0.3737 & \textbf{0.4872} & 0.2109 \\ 
		XGBoost Top8 & \textbf{0.8713} & \textbf{1.1968} & \textbf{0.3738} & 0.4769 & \textbf{0.2105} \\ 
		\bottomrule
	\end{tabular}
\end{table}

%\pagebreak
The top eight variables are now described and explained in more detail, with Section~\ref{sec:covariate} focusing on the (classic) covariate data, Section~\ref{sec:historic} explaining the \textit{HistAbility}, Section~\ref{sec:bookmaker:data} describing the \textit{logability}, and finally, Section~\ref{sec:pm:data} elaborates on the \textit{ave.PM} variable.

\subsection{Covariate data}\label{sec:covariate}

Similar to \cite{groll:21}, now we describe the first two types of data containing the match results of all UEFA EUROs 2004-2020 linked with the (classical) covariate data for each team. Therefore, the following five variables, which were chosen based on the variable importance as described before, are collected in advance of every UEFA EURO for each team:
	
\begin{description}
	\item[\textbf{\textit{GDP per capita:}}] \hfill \\ 
	The gross domestic product (GDP) per capita of each country is obtained. To accommodate the rise in GDP over the years, the logarithmized GDP per capita is used. This ensures comparability of the GDP over the different UEFA EUROs. Moreover, the most recent GDP value is used, which can slightly differ from the actual year of the corresponding UEFA EURO (source: \url{http://data.worldbank.org/indicator/NY.GDP.PCAP.CD}).
	\item[\textbf{\textit{Market value:}}] \hfill \\
	For each participating team, the market values are obtained from the website \url{http://www.transfermarkt.de}\footnote{As the first market values were published on this website shortly after the UEFA EURO 2004, the ones used for this tournament had to be estimated rather roughly.}. Additionally, the average market value is then calculated and logarithmized for each team, to secure comparability among the UEFA EUROs.
	\item[\textbf{\textit{FIFA ranking:}}] \hfill \\
	The official FIFA ranks are obtained for each team, reflecting the teams' performance in the last years.\footnote{The FIFA introduced a new system (called ``SUM") in 2018 for the calculation of the FIFA points, which determine the FIFA rank for each team. For more information on the method see \url{https://inside.fifa.com/fifa-world-ranking/procedure-men}.}
	\item[\textbf{\textit{UEFA points:}}] \hfill \\
	The UEFA points (officially named ``associations' club coefficients") determine a ranking across all UEFA members. For this purpose, the results of domestic clubs participating in UEFA competitions over the last five years are utilized, to calculate the UEFA points of a nation\footnote{Further information on how the points and ranking is computed can be found at the official UEFA website \url{https://www.uefa.com/nationalassociations/uefarankings/country/?year=2025}.}.
	\item[\textbf{\textit{Number of Champions league players:}}] \hfill \\
	For each national team the number of players who play at domestic clubs that reached the semi final in the preceding UEFA Champions League season is counted.\footnote{Sources: \url{http://kicker.de} and \url{http://transfermarkt.de}.} Note that since UEFA EURO 2020 the squad size has been increased to 26 players. To make this covariate comparable across all tournaments, it is normalized to a 23-player squad.
\end{description}
The procedure of collecting the covariate data was completed right after the final squads of each national team were announced, which typically occurs about a week before the start of the tournament. The five selected (classical) covariates also provide different views on the participating teams. With the GDP a purely economic factor is included, which has no direct sporting background. On the other hand, sporting factors are considered (market value, FIFA ranking, UEFA points), which are based on the performances of the players/teams. Finally, the number of Champions League players is a covariate providing both information for the team structure, but also to some extent for the teams' strength.

In addition to these five covariates, there are three enhanced variables derived from different ranking methods, which are described in more detail in the following sections. The final dataset is constructed in the same manner as in \cite{groll:21}, using the match results from past UEFA EUROs (Table~\ref{tab:results}) together with (in total eight) covariates for each team at the corresponding UEFA EURO (Table~\ref{tab:covar}).

\begin{table}[h]
	\small
	\caption{\label{data1} The following tables illustrate the underlying datasets, exemplary for the UEFA EURO 2020: (a) the results of four matches from the UEFA EURO 2020 and (b) some covariates for the corresponding teams.}
	\centering
	\subfloat[Table of results \label{tab:results}]{
	\begin{tabular}{lcr}
		\hline
		&  &  \\ 
		\hline
		HUN \includegraphics[width=0.4cm]{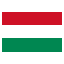} & 0-3 &  \includegraphics[width=0.4cm]{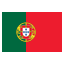} \;POR\\
		FRA\, \includegraphics[width=0.4cm]{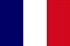} & 1-0 &  \includegraphics[width=0.4cm]{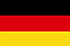} \;GER\\
		HUN \includegraphics[width=0.4cm]{HUN.png} & 1-1 &\includegraphics[width=0.4cm]{FRA.png} \;\,FRA\\
		POR \includegraphics[width=0.4cm]{POR.png} & 2-4 &  \includegraphics[width=0.4cm]{GER.png} \;GER\\
		\vdots & \vdots & \vdots  \\
		\hline
	\end{tabular}}
	\hspace*{0.3cm}
	\subfloat[Table of covariates \label{tab:covar}]{
	\begin{tabular}{llrrrrrr}
		\hline
		EURO & Team &  HistAbility & logability &  ave.PM & FIFA.rank & \ldots \\ 
		\hline
		2020 & Portugal  & $0.258$  &  $0.244$ & $0.110$ & $5$ & \ldots \\ 
		2020 &  Hungary & $-0.295$  &  $-0.202$ & $-0.010$ & $37$ & \ldots\\ 
		2020 &  France & $0.275$  & $0.334$ & $0.129$ & $2$ & \ldots \\ 
		2020 &  Germany & $0.225$  & $0.249$ & $0.170$ & $12$ & \ldots\\ 
		\vdots & \vdots & \vdots & \vdots & \vdots  &   \vdots  & $\ddots$ \\
		\hline
	\end{tabular}
	}
\end{table}

Identically to \cite{groll:21}, the target variable for all models, which will be introduced in Section~\ref{sec:methods}, are the number of goals a team scored in a match after 90 minutes. Therefore, a match is divided into two separate observations, each representing one team and their goals scored in this match. For each observation, we calculate the covariate difference for the two competing teams from the perspective of the first-mentioned team. For instance, Table~\ref{data2} shows parts of the final dataset used for the models, exemplary for the matches listed in Table~\ref{data1}.

\begin{table}[!h]
\small
\centering
\caption{Parts of the final dataset used for our models, including two observations for each match and feature differences between the participating teams for the respective covariates.}\label{data2}
\begin{tabular}{rllrrrrrr}
  \hline
Goals & Team & Opponent & Historic match & Bookmaker & average PM &  FIFA rank &  ... \\ 
 &  &  & abilities & abilities & player ranking &  &  ... \\ 
  \hline
    0 & Hungary & Portugal & -0.554 & -0.446 & -0.120 & 32 &  ...  \\ 
    3 & Portugal & Hungary & 0.554 & 0.446 & 0.120 & -32 &  ...  \\
    1 & France & Germany & 0.050 & 0.085 & -0.041 & -10 &  ...  \\
    0 & Germany & France & -0.050 & -0.085 & 0.041 & 10 &  ...  \\ 
    1 & Hungary & France & -0.570 & -0.536 & -0.139 & 35 &  ...  \\ 
    1 & France & Hungary & 0.570 & 0.536 & 0.139 & -35 &  ...  \\ 
    2 & Portugal & Germany & 0.033 & -0.005 & -0.060 & -7 &  ...  \\
    4 & Germany & Portugal & -0.033 & 0.005 & 0.060 & 7 &  ...  \\
	 \vdots & \vdots & \vdots & \vdots & \vdots & \vdots & \vdots & \vdots &  $\ddots$ \\
   \hline
\end{tabular}
\end{table}

\subsection{Current ability ranking based on historic matches}\label{sec:historic}

%\textbf{Disclaimer:} The following section was produced in collaboration with Achim Zeileis and Christophe Ley, as they provided this ability ranking, and is based on the one described in \cite{groll:21}.

The historic abilities can be estimated using Poisson models. \cite{LeyWieEet2018} investigated the performance of ten different models, with the bivariate Poisson model showing the best performance, closely followed by the independent Poisson model, which we will use here. It is based on all international matches played in the last eight years preceding the corresponding UEFA EURO. For each match the competing teams, result, and match venue (time, place, match importance) are needed, as they decide on the weights used in the ranking method. As a short overview, Table~\ref{tab:historicdata} shows an excerpt of the underlying dataset, exemplary for the UEFA EURO 2020\footnote{Note that the UEFA EURO 2020 took place in summer 2021 due to the COVID-19 pandemic.}.

\begin{table}[ht]
	\caption{Historical match data used to estimate team abilities, exemplary for matches shortly before the UEFA EURO 2020.}
	\label{tab:historicdata}
	\centering
	\begin{tabular}{lllcrrr}
		\hline
		Date & Home team & Away team & Score & Country & Neutral & Match type\\
		\hline
		2021-06-02 & England & Austria & 1-0 & England & no & friendly\\
		2021-06-02 & France & Wales & 3-0 & France & no & friendly\\
		2021-06-02 & Germany & Denmark & 1-1 & Austria & yes & friendly\\
		2021-06-02 & Netherlands & Scotland & 2-2 & Portugal & yes & friendly\\
		\vdots & \vdots & \vdots & \vdots & \vdots & \vdots & \vdots \\
		\hline
	\end{tabular}
\end{table}
For two competing teams~$i,j \hspace{0.2cm} (i \neq j)$ in match~$m$, we model their number of goals via random variables $G_{i,m}$ and $G_{j,m}$. These are assumed to be independently Poisson distributed, resulting in the following probability function:
$$
{\rm P}(G_{i,m}=x, G_{j,m}=y) = 
\frac{\lambda_{i,m}^x}{x!} \exp(-\lambda_{i,m})\cdot
\frac{\lambda_{j,m}^y}{y!} \exp(-\lambda_{j,m}), 
$$
with $\lambda_{i,m}$ and $\lambda_{j,m}$ being the expected number of goals for the competing teams. These parameters are specified in the following way
\begin{eqnarray} \label{eqn:1}
	\label{independentpoisson1}\lambda_{i,m}&=&\exp\big(\beta_0 + (r_{i}-r_{j})+h_i\cdot \mathds{1}(\mbox{team $i$ playing at home}\big), \\
	\label{independentpoisson2}\lambda_{j,m}&=&\exp\big(\beta_0 + (r_{j}-r_{i})+h_j\cdot \mathds{1}(\mbox{team $j$ playing at home}\big),
\end{eqnarray}
with $\beta_0\in\mathds{R}$ being a standard intercept, $r_i, r_j \in\mathds{R}$ being the abilities of the two competing teams, and $h_i, h_j \in \mathds{R}$ reflecting possible (team-specific) home advantage effects. Additionally, for identifiability purposes, we add the constraint $\sum_i r_i = 0$. Under the independence assumption, we get the following joint likelihood function:
\begin{equation*}
	L = \prod_{m=1}^{M}\left({\rm P}(G_{i,m}=g_{i,m}, G_{j,m}=g_{j,m})\right)^{w_{m}}, 
\end{equation*}
with $g_{i,m}$ and $g_{j,m}$ representing the true score by teams~$i$ and $j$ in match~$m$, and weight $w_{m}$ being the product of the two weights $w_{time,m}(x_m)$ (for time decay) and $w_{type,m}$ (for match importance). These weights are introduced in \cite{LeyWieEet2018} and ensure that recent matches and matches with higher relevance are given more weight.

\begin{figure}[!ht]
	\centering
	\includegraphics[width=0.5\textwidth]{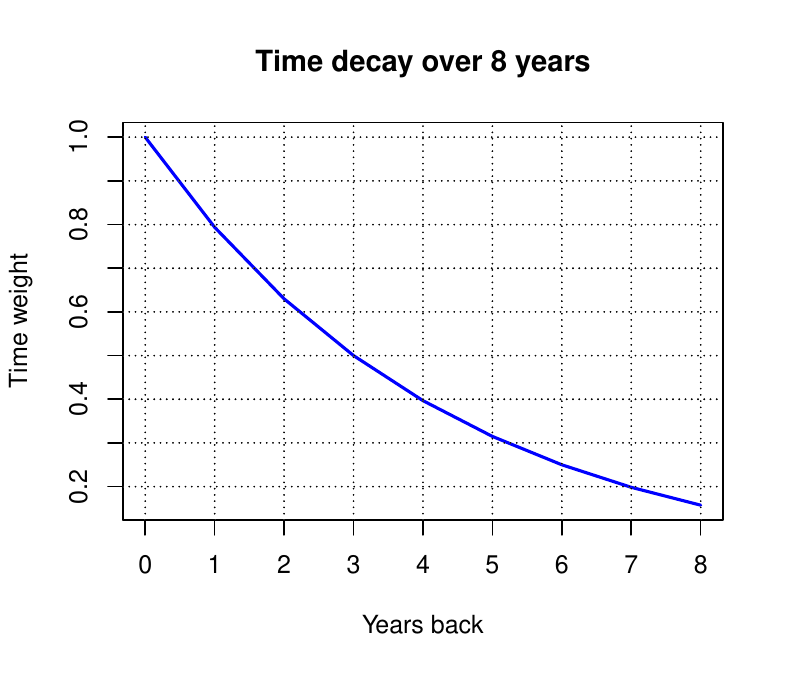}
	\caption{Plot showing the time weight of a match played $x_m$ days ago obtained through the $w_{time,m}(x_m)$ function for a \emph{Half period} of three years (i.e. $3*365.25 = 1095.75$ days).}
	\label{timeweight}
\end{figure}

For the time decay a continuous function is used, assigning a match $m \in \{1,...,M\}$, which is played $x_m$ days back, a weight of
\begin{equation*}\label{smoother}
	w_{time,m}(x_m) = \left(\frac{1}{2}\right)^{\frac{x_m}{\mbox{\small Half period}}}\,.
\end{equation*}
This way, a match played today gets weight one, while a match played \emph{Half period} days ago gets a weight of one half. \cite{LeyWieEet2018} tested several \emph{Half periods} (from half a year to six years), with a \emph{Half period} of three years (i.e. $3 \cdot 365.25 = 1095.75$ days) yielding the best Rank Probability Score (RPS; \citealp{Epstein:69}). Figure~\ref{timeweight} shows the resulting time weights for a \emph{Half period} of three years.

Moreover, for the match importance, match $m$ is assigned with the following weight regarding the match type:
\begin{equation*}\label{match:importance}
	w_{type,m} = 
	\begin{cases} 
		4 & \text{World Cup match}, \\
		3 & \text{confederation tournament match}, \\
		2.5 & \text{confederation or World Cup qualifier match}, \\
		1 & \text{otherwise (e.g.\ friendly matches, etc.)}. \\
	\end{cases}
\end{equation*}
Finally, the values of the strength parameters $r_1,\ldots,r_n$, which are the main parameters of interest and in the following will serve as an enhanced variable, are calculated via maximum likelihood estimation using the historic match data as described before. Furthermore, note that principally these strength parameters can also be used to directly predict upcoming matches via Equation~\eqref{independentpoisson1} and \eqref{independentpoisson2}.

\subsection{Bookmaker consensus model}\label{sec:bookmaker:data}

%\textbf{Disclaimer:} The following section was produced in collaboration with Achim Zeileis, as he provided the bookmaker consensus model for us, and is based on the one described in \cite{groll:21}.

The bookmaker consensus model was introduced by \citet{Leit:2010a}, with the aim of estimating team abilities from bookmaker winning odds, which reflect the bookmakers' ``expert" knowledge. Therefore, the bookmakers' odds need to be collected for each UEFA EURO from 2004-2024. To estimate the team abilities for the UEFA EUROs 2004-2020, which are needed for our training data, the odds were obtained from previous works on the UEFA EUROs 2008, 2012, 2016, and 2020 \citep{Leit:2010a,Zeil:2012,Zeil:2016,groll:21}. For these tournaments, several odds from different bookmakers were collected, while for the UEFA EURO 2004 only the odds from ODDSET\footnote{ODDSET is a German betting agency, which first offered this type of bet at the UEFA EURO 2004.} could be obtained (see Table~\ref{tab:odds2004} in Appendix~\ref{sec:appendix:ata}). To estimate the team abilities for the current UEFA EURO 2024, each team's winning odds were collected from 28 bookmakers via the websites \url{https://www.oddschecker.com/} and \url{https://www.bwin.com/} (see Table~\ref{tab:odds2020} in Appendix~\ref{sec:appendix:ata}).		

Since bookmakers are profit-oriented companies, these odds are not totally fair and therefore have to be adjusted to (a) remove the profit margin (so-called ``overround") and (b) to account for the invested stake \citep[for more information see][]{ref:Henery:1999, ref:Forrest+Goddard+Simmons:2005}. Therefore, the quoted odds for team~$i$ are obtained from fair odds using the following relation:
$$
\mbox{\it quoted odds}_i = \mbox{\it odds}_i \cdot \delta + 1,
$$
where the stake is accounted for by the $+1$, reflecting the bet investment, which must be returned in addition to the net winnings in case of a win. The bookmaker only pays out the share $\delta < 1$ of the bet, while the share $1-\delta$ is the bookmakers' profit. Similar to \cite{Leit:2010a}, we also assume a constant overround $\delta$ for all teams to win the tournament, resulting in a median value of $16.8\%$ across all bookmakers. 

The ``cleaned" odds from the 28 bookmakers are then scaled using logarithmic transformation and averaged. The resulting log-odds ($l_i$) for team~$i$, averaged across all bookmakers, are then used to compute winning probabilities $p_i$ for team~$i$ via
$$
p_i = \frac{1}{\exp(l_i) + 1}.
$$
These probabilities then serve as basis to estimate team abilities. Therefore, the following strategy (so-called ``inverse" tournament simulation) can be used:
\begin{enumerate}
	\item If team abilities are available (along with the offset $\beta_0$ and the home advantage parameter $h$), then the independent Poisson model from Equation~\eqref{eqn:1} can be used to simulate every possible match.
	\item Given the simulated matches, the entire tournament can be simulated to determine the tournament winner.
	\item To get proper winning probabilities for each team, this tournament simulation is executed 100,000 times.
\end{enumerate}
Here, we set the offset to $\beta_0 = 0.15$ and drop the home advantage, i.e., set $h = 0$. These values are set because (a) they are not too far from empirical results for recent football championships and (b) the precise values have only minor influence on the main quantities of interest, the team strength parameters.

These are found through the iterative method used in \cite{Leit:2010a}, which uses the following procedure:
\begin{enumerate}
	\item Simulate the tournament as described before using some initial ability specification (e.g.\ based on the bookmakers average log-odds, i.e.\ $abilities_{iter}, iter =~1$).
	\item Derive the simulated winning probabilities $\tilde{p}_i$ and transform them to log-odds scale $\tilde{l}_i$.
	\item Compute the loss between the simulated log-odds and bookmaker log-odds $l_i$ over all participating teams $i\in\{1,\ldots,N\}$: $\text{loss}_{\text{iter}} = \sqrt{\frac{1}{N} \sum_{i=1}^{N} (\tilde{l}_i - l_i)^2}$.
	\item If $loss_{iter} < 0.05$ stop the procedure.
	\item Adjust the abilities for each team i: $abilities_{iter+1} = abilities_{iter} - \frac{(\tilde{l}_i - l_i)}{|\tilde{l}_i - l_i| \cdot \frac{\texttt{0.01}}{\texttt{iter}^{\texttt{0.1}}} },$ and increment $iter$ by 1.
\end{enumerate}
This whole procedure is repeated, until step~4 breaks the process. The final log-abilities are centered around zero as their mean is not identified.
These team abilities (\textit{logability}) for each team then serve as additional enhanced covariate for our models.

\subsection{Plus-minus player rating}\label{sec:pm:data}

%\textbf{Disclaimer:} The following section was produced in collaboration with Lars M. Hvattum, as he provided the plus-minus player rating for us, and is based on the one described in \cite{groll:21}.

The plus-minus (PM) player ratings, proposed by \citet{Hv19}, are based on historic matches, but in contrast to the ability ranking from Section~\ref{sec:historic}, these are additionally also based on club matches. Moreover, information on the starting line-ups, goals scored, substitutions and red cards are needed for each match. Therefore, for each event the time-stamp is needed and for substitutions and red cards also the corresponding player is required. Then each match is divided into segments, with a new segment starting, whenever the current set of players playing changes (through substitution or red card). For a short overview on the fundamental dataset, see Tables~\ref{tab:pmdata:1} and \ref{tab:pmdata:2}.

\begin{table}[htbp]
	\centering
	\caption{Exemplary matches (from the year UEFA EURO 2020 took place) used for estimating plus-minus player ratings.}
	\label{tab:pmdata:1}
	\begin{tabular}{clll}
		\hline
		Date  & Home team & Away team & Neutral \\
		\hline
		2021-03-13 & Real Madrid & Elche & no \\
		$\vdots$ & $\vdots$ & $\vdots$ & $\vdots$ \\
		2021-06-02 & Romania & Georgia & no \\
		$\vdots$ & $\vdots$ & $\vdots$ & $\vdots$ \\
		\hline
	\end{tabular}%
	\label{tab:addlabel}%
\end{table}%

\begin{table}[htbp]
	\centering
	\caption{Extended dataset from the matches of Table~\ref{tab:pmdata:1} displaying the segments each match is divided into, with further information on the participating players, red cards at the beginning of a segment, and goals scored during a segment, both for the home team (HT) and away team (AT), respectively.}
	\label{tab:pmdata:2}
	\begin{tabular}{cllcccc}
		\hline
		&       &       & \multicolumn{2}{l}{Red cards} & \multicolumn{2}{l}{Goals} \\
		Time  & HT line-up & AT line-up & HT & AT & HT & AT \\
		\hline
		0--36  & Courtois, Ramos, ... & Badia, Verdu, ... & 0     & 0     & 0     & 0 \\
		36--61 & Courtois, Ramos, ... & Badia, Verdu, ... & 0     & 0     & 0     & 1 \\
		$\vdots$  &$\vdots$  &  $\vdots$ &  $\vdots$  & $\vdots$ & $\vdots$  & $\vdots$ \\
		0--42  & Vlad, Ganea, ... & Kupatadze, Chabradze, ... & 0     & 0     & 0     & 0 \\
		42--57 & Vlad, Ganea, ... & Kupatadze, Chabradze, ... & 1     & 0     & 0     & 0 \\
		$\vdots$  &$\vdots$  &  $\vdots$ &  $\vdots$  & $\vdots$ & $\vdots$  & $\vdots$ \\
		66--77 & Vlad, Ganea, ... & Kupatadze, Chabradze, ... & 1     & 0     & 0     & 1 \\
		$\vdots$  &$\vdots$  &  $\vdots$ &  $\vdots$  & $\vdots$ & $\vdots$  & $\vdots$ \\
		\hline
	\end{tabular}%
	\label{tab:addlabel}%
\end{table}%

Next, we explain the method that is used to compute the PM ratings. The fundamental idea of the rating is a simple linear regression
$$
y_i = \sum_{j} \beta_j x_{ij} + \epsilon_i,
$$
with $y_i$ being the goal difference for segment $i$ (from the home team's perspective), $\epsilon_i$ being an error term, and $x_{ij}$ representing if player $j$ is playing for the home team in segment $i$ ($x_{ij} = 1$), the away team ($x_{ij} = -1$), or currently is not playing ($x_{ij} = 0$). The estimated regression coefficients $\beta_j$ can then be understood as the player ratings. As this approach is too basic for football player ratings, some adjustments must be made (see \citealp{PaHv20}):
\begin{itemize}
	\item The home-field advantage is modelled using a single covariate for each competition.
	\item To handle segments with less players (due to red cards), additional covariates are included and the remaining player ratings are scaled.
	\item As players participate in different competitions, the players' ratings are adjusted based on the different competitions they participated in.
	\item Age adjustment variables are added for the players' ratings.
	\item Furthermore, each segment is weighted based on the time since the match was played, the duration of the segment, and the game state (e.g.\ goal difference)
	\item The ratings are estimated using ridge regression (see \citealp{HoeKen:70}).
	\item For coefficients corresponding to player ratings, the regularization is adjusted ad hoc, under the assumption that a player is more likely to be of similar playing strength as
		the most common teammates, rather than an average player.
		%For variables corresponding to the player rating a different regularization is used, with these variables beeing shrunk towards an average of the variables from a group of comparable players (e.g.\ players having 		played the most time together).
\end{itemize}

Based on the historic data, for each player a PM rating is calculated as described before, which will then be used to compute new covariates for the participating UEFA EURO teams:
\begin{itemize}
	\item A mean PM rating of the nominated squad,
	\item a median PM rating of the nominated squad,
	\item a mean PM rating of the best eleven nominated players, and
	\item the number of players missing in the squad that would be included in the best 11 players of the squad, and have played at least one international match in the last two years before the tournament.
\end{itemize}

For this project, we only focus on the mean PM (\textit{ave.PM}) team ratings. As already explained in Section~\ref{sec:data}, our analysis indicates that the \textit{ave.PM} player rating is more relevant compared to other PM ranking variants. Therefore, we only included the \textit{ave.PM} player rating in our dataset for the machine learning models, which will be introduced in Section~\ref{sec:methods}. Additionally, also \cite{groll:21} showed that the \textit{ave.PM} player rating is most important among the different PM variants considered for modeling purposes.

\subsection{Combine enhanced variables with covariate data}

As mentioned earlier, the three enhanced variables derived from the ranking methods described before, are added to the covariate data for the corresponding UEFA EURO, serving as new (highly informative) covariates (similar to \citealp{groll:21}). Therefore, we estimate the $HistAbility$ for every team based on their historic matches as a new covariate (see Section~\ref{sec:historic}). Additionally, we compute the \textit{logability} for each team, based on the bookmakers winning odds (see Section~\ref{sec:bookmaker:data}). Finally, the \textit{ave.PM} covariate is also derived from historic match data, serving as another covariate for each team (see Section~\ref{sec:pm:data}). All of the three enhanced variables are calculated shortly before the start of the corresponding UEFA EURO.

%%%%%%%%%%%%%%%%%%%%%%%%%%%%
%%%%%%%%%%%%%%%%%%%%%%%%%%%%
\section{Machine learning models}\label{sec:methods}

In this section, we describe the fundamentals of four models used for predicting football matches. We start with three classical modeling approaches: LASSO-regularized Poisson regression, random forests, and extreme gradient boosting. Finally, we introduce a novel combined model that employs the predictions of all three classical models.

%%%%%%%%%%%%%%%%%%%%%%%%%%%%
\subsection{LASSO}\label{subsec:lasso}
%\red{[AG: This whole section might be too long and detailed; also, there is a little bit of a mess regarding the indices $i$ and $j$ for the 2 teams and then later on the log-likelihood would sum over all observations, but there is currently no proper index for the individual observations; but before spending some effort on fixing this, it wanted to await your input as those parts might be remove anyway]}
The following LASSO model is based on a Poisson generalized linear model (GLM; \citealp{Fahrmeir2013}), and regularized by the so-called least absolute shrinkage and selection operator (LASSO), proposed by \cite{Tibshirani:96}. Versions of this model were also used in past football tournament predictions \citep{GroSchTut:2015,GroAbe:2013}, and evaluated by \cite{SchauGroll2018,groll:21}.
As we assume a Poisson-distributed response, the corresponding discrete density is given by
$$
f(y_{ij}|\mathbf{x}_{ij}) = \frac{\lambda_{ij}^{y_{ij}}exp(-\lambda_{ij})}{y_{ij}!}, \qquad E[y_{ij}|\mathbf{x}_{ij}] = \lambda_{ij},
$$
with $\mathbf{x}_{ij} = (1, x_{ij,1},\ldots,x_{ij,p})^\top \in \mathds{R}^{p+1}$ being the $p+1$-dimensional vector of $p$ feature differences of two competing teams $i$ and $j$ from the perspective of the first-named team $i$, and $y_{ij}$ being the number of goals scored by team $i$ in a match against team $j$ %$i \in \{1,...,n\}$ 
(see Table~\ref{data2}). Note that ${x}_{ij,0} = 1$ is accounting for the intercept.
To model the parameter $\lambda_{ij}$, which reflects the expected number of goals, we use the linear predictor $\eta_{ij} = \mathbf{x}_{ij}^\top \pmb{\beta}$ and apply the exponential response function to guarantee $\lambda_{ij} > 0 $ (see \citealp{Fahrmeir2013}), i.e.
$$
\lambda_{ij} = \exp(\eta_{ij}) = \exp(\mathbf{x}_{ij}^\top \pmb{\beta}).
$$
Moreover, the parameter vector $\pmb{\beta}$ determines the influence of each feature difference on $\lambda_{ij}$, including the intercept $\beta_0$, which can be understood as the average number of goals (ceteris paribus).
The density, also known as the likelihood $L_{ij}(\pmb{\beta})$, depends on $\pmb{\beta}$ through $\lambda_{ij}$.
Assuming conditional independence between all matches and observations, the joint likelihood is the product of the single likelihoods, i.e.:
$$
L(\pmb{\beta}) = \prod_{i,j} L_{ij}(\pmb{\beta}).
$$
The log-likelihood is then given by:
\begin{eqnarray}
	l(\pmb{\beta}) = \log \Big( \prod_{i,j} L_{ij}(\pmb{\beta}) \Big) = \sum_{i,j} \log \big(L_{i,j}(\pmb{\beta})\big) &&= \sum_{i,j} \Big(y_{ij} \log(\lambda_{ij}) - \lambda_{ij} - \log(y_{ij}!)\Big) \nonumber \\
	&&= \sum_{i,j} \Big(y_{ij} \log(\lambda_{ij}) - \lambda_{ij}\Big), \nonumber
\end{eqnarray}
where the constant factor $- \log(y_{ij}!)$ can be ignored, as it is independent of $\pmb{\beta}$ \citep{Fahrmeir2013}.
To introduce regularization and estimate $\pmb{\beta}$, we maximize the penalized log-likelihood (see also \citealp{SchauGroll2018,GroSchTut:2015}):
\begin{eqnarray}\label{eq}
	l_p(\pmb{\beta}) = l(\pmb{\beta}) - \xi P(\pmb{\beta}), \nonumber
\end{eqnarray}
where $P(\pmb{\beta})$ represents the penalty term. There are two frequently used options for this penalty, namely a $L_1$-penalty (LASSO) or $L_2$-penalty (ridge penalty; \citealp{HoeKen:70}). As LASSO is able to set parameters $\beta_k$ to zero, we use the LASSO penalty $P(\pmb{\beta}) = \sum_{k=1}^{p} |\beta_k|$ \citep{Tibshirani:96}. Moreover, this means that some covariates may not be included in the model, which indicates that LASSO basically performs variable selection.

The tuning parameter $\xi$ is obtained using \texttt{cv.glmnet} function from the \texttt{R} package \texttt{glmnet}  \citep{FrieEtAl:2010}. This executes 10-fold cross-validation, where the $\xi $ value is chosen that minimizes the cross-validation error.
For a new observation, the covariates are plugged into the linear predictor, resulting in a prediction for the intensity parameter $\lambda$ of a Poisson distribution.

%\end{redsection}
%%%%%%%%%%%%%%%%%%%%%%%%%%%%
\subsection{Random forests}\label{subsec:forest}
% \begin{itemize}
% \item Explain regression trees and forests on goals in more detail
% \item Mention that other types of forests for football data could be used as well; see Stat. Modelling
% \item But \texttt{party} regression forest performed very satisfactory and is used in the following
% \item Show plot of how tree generally works and maybe variable importance (both without abilities)
% \end{itemize}
%The following section is based on the equivalent one from \cite{groll:21}.

\emph{Random forests} are an ensemble of many (e.g.\ $T = 5000$) regression trees, initially proposed by \citet{Breiman:2001}. The aim of a single tree is to recursively divide the dataset into (mostly two) partitions, such that within each partition the dataset is homogeneous regarding the response variable, and between the different partitions the dataset is heterogeneous. As already mentioned earlier, in our case the response variable is the number of goals from a team.

The advantage of random forests lies in the (parallel) combination of many trees, leading to a reduced variance of the predictions, while still maintaining the merit of unbiasedness. For this reason, the random forest introduces two different sources of randomness. First of all, each tree in the random forest is constructed using a bootstrap sample from the training data (usually 63.2\%). Additionally, for a split in the tree, only a subset of \texttt{mtry} features is randomly sampled as candidates to find the best split, with \texttt{mtry} being the most important tuning parameter. For a more detailed description on how random forests can be used to predict football tournaments, see also \cite{GroEtAl:WM2018b}.

In our case, for the prediction of a new observation $i$, the covariates are dropped down each tree resulting in \( T \) predictions, which are then averaged:
$$
\hat{y}_i = \frac{1}{T} \sum_{t=1}^{T} \hat{y}_i^{(t)},
$$
with $\hat{y}_i^{(t)}$ being the prediction of the $t$-th tree.

We use a special random forest implementation, called \texttt{cforest} from the \texttt{party} package \citep{Hotetal:2006}. This random forest was advantageous over the classical random forest implementation \citep{Breiman:2001} from the \texttt{R}-package \texttt{ranger} \citep{ranger} in past analysis of football tournaments, see e.g.\ \citet{SchauGroll2018}, \citet{GroEtAl:WM2018b} and \citet{groll:21}, which is why in the present manuscript only the \texttt{cforest} is evaluated.

The Cforest approach %\texttt{cforest} 
is based on conditional inference trees (ctrees), proposed by \cite{Hothorn+Hornik+Zeileis:2006}. Using ctrees instead of classical regression trees has the advantage of unbiasedness in the selection process of a split variable, when the covariates have different scales (e.g.\ FIFA ranking vs. market value). This feature also holds for Cforest, %the \texttt{cforest}, 
as it is constructed using ctrees.

\begin{figure}[!ht]
	\centering
	\includegraphics[width=\textwidth,height=\textheight,keepaspectratio]{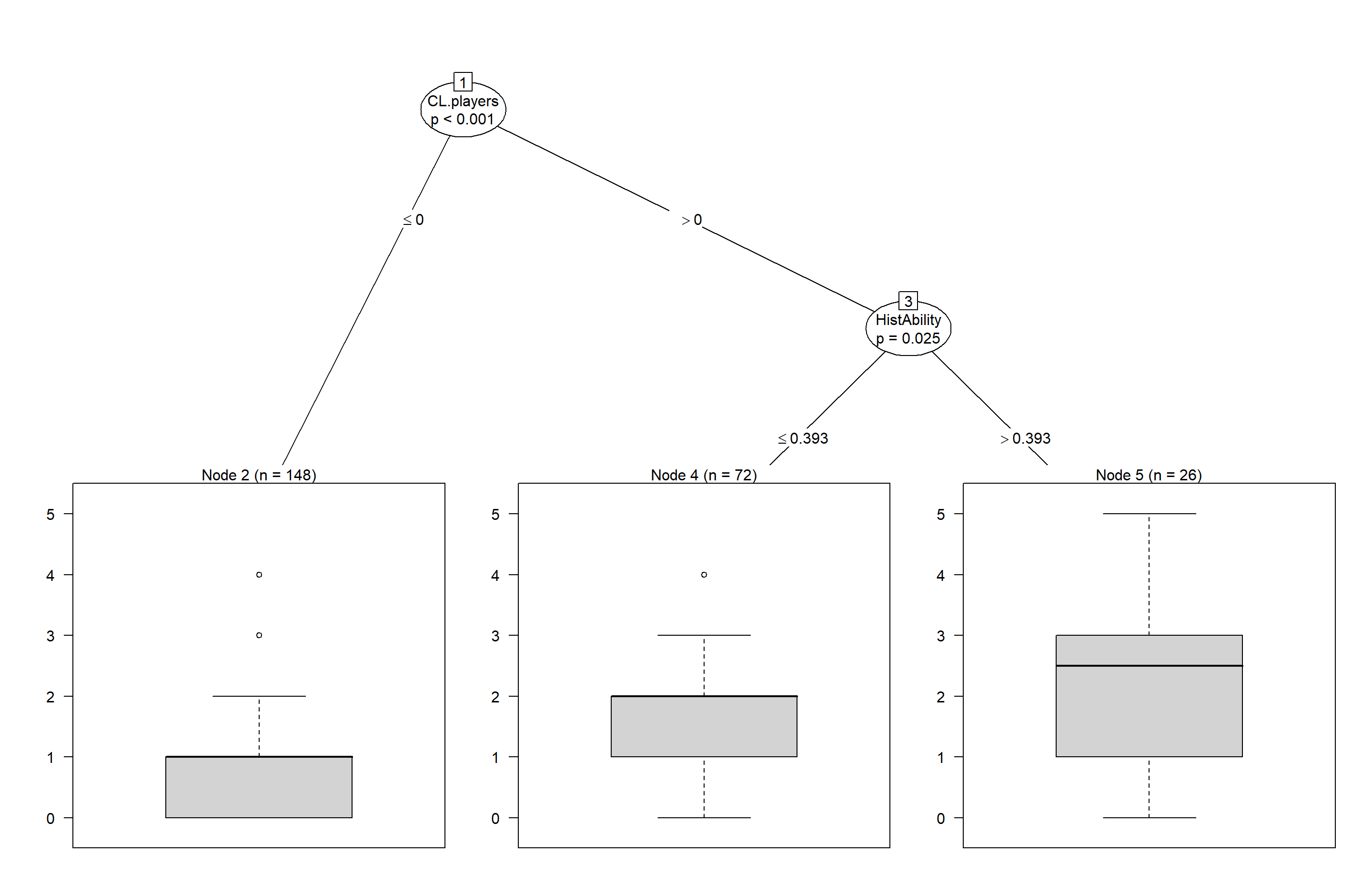}
	\caption{Exemplary \texttt{ctree} created on one single bootstrap sample of our training data, i.e.\ all matches from the  EUROs 2004-2020.}
	\label{ctree}
\end{figure}

For illustration, Figure~\ref{ctree} shows an exemplary ctree created from a bootstrap sample of our training data (i.e.\ all matches from the  EUROs 2004-2020), resulting in three different (more or less) homogenous partitions.

%\pagebreak

%%%%%%%%%%%%%%%%%%%%%%%%%%%%
\subsection{Extreme gradient boosting}\label{subsec:xgboost}
%The following section is based on the equivalent one from \cite{groll:21}.

In contrast to the parallel ensemble methods, like the random forests described before, extreme gradient boosting is a sequential ensemble method, referring to the class of \emph{boosting} algorithms. This class originates from the machine learning community \citep{FreuScha:96}, with first boosting algorithms proposed by \cite{Schapire:90} and \cite{Freund:95}. The basic idea of this concept is to sequentially combine multiple weak learners, resulting in a strong learner with better performance.

This method was then transferred from the machine learning to a statistical perspective \citep{FriHasTib:2000, Fri:2001}. For example, \cite{MayrEtAl:pt1:2014} describe this evolution of boosting, naming several advantages such as that statistical boosting algorithms are interpretable, stable for high-dimensional data and also can implicitly incorporate variable selection in the fitting process. 

As the learners can be based on various different models, \cite{Fri:2001} introduced the idea of using regression trees as weak learners. Therefore, each tree is trained on the errors left by the preceding tree and added (down-scaled by a learning rate) to the overall model, to further improve it. %the complete model.

\cite{chen2016xgboost} further improved tree-based boosting by adding regularization terms that penalize the complexity of the regression tree, which helps to avoid overfitting, showing success in various machine learning challenges. This technique is called \emph{extreme gradient boosting} (XGBoost) and is implemented in the \texttt{xgb.train} function from the \texttt{xgboost} R package \citep{chen:2024}.

For a given dataset with features $\xb_i \in \mathds{R}^p$ the XGBoost incorporates $K$ additive trees to predict the expected number of goals $y_i \in \mathds{R}$ for observation $i \in \{1,\ldots,n\}$, i.e.:
$$
\hat{y}_i = \sum_{k=1}^{K} f_k(\xb_i), \quad f_k \in F,
$$
where $ F = \{f(x) = w_{q(x)} \}$ is the set of trees, $f_k$ corresponds to a tree structure $q : \mathds{R}^p \rightarrow T$ and leaf weights $\wb \in \mathds{R}^T$, with $w_i$ representing the score on the $i$-th leaf, and $T$ is the total number of leaves in each tree. %\citep{chen2016xgboost}.
Note that in our case $\xb_i$ is already the variable difference of two competing teams.

To determine the functions that are part of the model, we minimize the following objective:
$$
L(\phi) = \sum_i l(\hat{y}_i,y_i) + \sum_k \Omega(f_k), \\
$$
where $\Omega(f) = \gamma\cdot T + \frac{1}{2} \lambda \|\wb\|^2$ penalizes the complexity of the tree, and $l(\hat{y}_i,y_i)$ is the loss between the models prediction $\hat{y}_i$ and actual value $y_i$. %\citep{chen2016xgboost}.

To further prevent overfitting, the (typically small) shrinkage parameter $\eta$ is introduced, which scales newly added weights after each step of tree boosting. %\citep{chen2016xgboost}. 
This parameter can be interpreted as a learning rate. Overall, we obtain the final prediction
$$
\hat{y}_i = \sum_{k=1}^{K} \eta f_k(x_i).
$$

The challenge of the XGBoost are the several tuning parameters involved, e.g.\ the learning rate, number of boosting iterations and additional regularization parameters. Therefore, instead of tuning it, we decided to fix the learning rate to a reasonable value ($\eta = 0.1$), reducing the complexity of the tuning process. To obtain values for the other tuning parameters, we used multivariate 10-fold cross-validation via the \texttt{xgb.cv} function for specified parameter grids.

\subsection{Combined model}
As all three models explained before rely on different approaches to predict the mean value for the number of goals a team scores in a match of a UEFA EURO tournament, we will now linearly combine the three individual predictions, i.e.\ $f_{LASSO}, f_{Cforest}, f_{XGBoost}$, to get a new combined prediction. The idea is that this approach combines all advantages of the three models to improve the prediction compared to the single models. To combine the predictions, we introduce weights $\{w_1,w_2,w_3\}$ for each model's prediction, which indicate the share a model contributes to the combined prediction. The final prediction of the mean value for the number of goals $\hat{y}_i$ then looks like this:
\begin{align}
	\hat{y}_i &= w_1 * f_{LASSO} &&+\, w_2 * f_{Cforest} &&+\, w_3 * f_{XGBoost} \nonumber \\
	&= w_1 * \Big( exp(\beta_0 + x_i \beta) \Big) &&+\, w_2 * \Big( \frac{1}{B} \sum_{b=1}^{B} \hat{y}_i^{(b)} \Big) &&+\, w_3 * \Big( \sum_{k=1}^{K} \eta f_k(x_i) \Big), \nonumber
\end{align}
under the side constraint $\sum_{j} w_j = 1$, to get realistic values for the mean number of goals and to not over- or underestimate the number of goals. The weights for the combined model are obtained by a tuning-function, which is described later in Section~\ref{model:tuning}.
\\
\\
As the predictions of all models are not integers they can not be directly used for predicting a single match. Therefore, we follow \cite{groll:21} in using a Poisson distribution $Po(\lambda)$ to randomly sample the number of goals a team scores, with the parameter $\lambda$ being predicted by our models. The result of a specific match can then be predicted using two (conditionally) independent Poisson distributions for each team. Assuming Poisson-distributed number of goals, we can also calculate probabilities for the three possible match outcomes, which will be explained in Section~\ref{sec:combine}.

%%%%%%%%%%%%%%%%%%%%%%%%%%%%
\section{Model performance}\label{sec:combine}
%%%%%%%%%%%%%%%%%%%%%%%%%%%%
In the following, we closely follow \citet{groll:21} and investigate the predictive performance of the proposed combined model, while comparing it also to the performance of the single models (LASSO, Cforest, XGBoost). For that reason, each model is evaluated using a tournament-specific cross validation on the UEFA EUROs 2004-2020 data. The following strategy guarantees that each match is once used for an out-of-sample prediction: 
\begin{enumerate}
	\item Create a training dataset by selecting four out of the five UEFA EUROs; the remaining UEFA EURO will serve as the test data. \vspace{0.1cm}
	\item Tune each method on this training dataset.\vspace{0.1cm}
	\item Fit each method (with optimal tuning parameters) to this training dataset. \vspace{0.1cm}
	\item Forecast the outcomes of the remaining UEFA EURO using each method. \vspace{0.1cm}
	\item Repeat steps 1-4, ensuring that each UEFA EURO is used as test data once. \vspace{0.1cm}
	\item Evaluate and compare the predicted outcomes with the actual results for all methods using different performance measures.
\end{enumerate}

For evaluating the predictions, predicted probabilities $\hat\pi_{1i},\hat\pi_{2i},\hat\pi_{3i},~i=1,\ldots,N$ are computed for the three ordinal match outcomes. Therefore, let $G_{1i}\sim Po(\hat\lambda_{1i})$ and $G_{2i}\sim Po(\hat\lambda_{2i})$ denote Poisson-distributed random variables for two competing teams, reflecting the number of goals scored by each of them in match~$i=1,\ldots,N$. The parameters $\hat\lambda_{1i}$ and $\hat\lambda_{2i}$ represent the estimates received from our models. Assuming this setup, the predicted probabilities $\hat \pi_{1i}=P(G_{1i}>G_{2i}), \hat \pi_{2i}=P(G_{1i}=G_{2i})$ and $\hat \pi_{3i}=P(G_{1i}<G_{2i})$ can be computed using the Skellam distribution (see Appendix~\ref{sec:general} for more details). Furthermore, it is necessary to know the actual match outcome (win home team, draw, win away team) for match~$i$, which is denoted by $\tilde y_i\in\{1,2,3\}$. These outcomes, along with the resulting predicted probabilities $\hat\pi_{1i},\hat\pi_{2i},\hat\pi_{3i}$, are then used to calculate the following performance measures:
\begin{itemize}
	\item {\it multinomial likelihood} (ML): It represents the probability of correct predictions, and for a individual match $i$ is defined as
	$$
	ML_i = \hat \pi_{1i}^{\delta_{1\tilde y_i}} \hat \pi_{2i}^{\delta_{2\tilde y_i}} \hat \pi_{3i}^{\delta_{3 \tilde y_i}},
	$$
	with  $\delta_{r\tilde y_i}$ denoting Kronecker's delta, which is defined as
	$$
	\delta_{ij} = \begin{cases}1,\quad \text{if}\,\, i=j\,,\\
		0,\quad \text{otherwise}\,.
	\end{cases}
	$$
	The overall ML for all predicted matches is then defined as the average of the single $ML_i$, where a larger value indicates a better prediction.\vspace{0.1cm}
	\item {\it classification rate} (CR): It reflects the proportion of correctly predicted match outcomes. To determine whether match~$i$ was correctly classified, the following indicator function is used:
	$$
	CR_i = \mathds{1}(\tilde y_i=\underset{r\in\{1,2,3\}}{\mbox{arg\,max }}(\hat\pi_{ri}))
	$$
	The CR for all predicted matches is the average of the single $CR_i$, with a larger value corresponding to better predictions.
	\item {\it rank probability score} (RPS): It is an error measure, taking the ordinal structure of the responses into account. For match~$i$, it is defined as
	$$RPS_i = \frac{1}{3-1} \sum\limits_{r=1}^{3-1}\left( \sum\limits_{l=1}^{r}(\hat\pi_{li} - \delta_{l\tilde y_i})\right)^{2},$$
	The RPS for all matches is defined as the average of the single $RPS_i$. In contrast to the other two measures, a low score reflects a good prediction.
\end{itemize}

As we are predicting the mean value of the number of goals a team scores, which is also crucial for simulating the UEFA EURO tournament (see Section~\ref{sec:prediction}), we also want to investigate the models performance regarding the number of goals. Therefore, we calculate the following absolute errors:
\[
\begin{aligned}
	MAE_{j,i} &= |y_{j,i} - \hat{y}_{j,i}|, \\
	MAE_{i}' &= \left| (y_{1,i} - y_{2,i}) - (\hat{y}_{1,i} - \hat{y}_{2,i}) \right|,
\end{aligned}
\]
where \( y_{j,i} \) denotes the actual number of goals team $j \in \{1,2\}$ (first- or second-named team) scored in match~$m$, and \( \hat{y}_{j,i} \) denotes the corresponding prediction from a model. 
To obtain the mean absolute errors, we average these values over all matches:
\[
\begin{aligned}
	MAE_{goals} &= \frac{1}{2N} \sum\limits_{i=1}^N\sum_{j\in\{1,2\}} MAE_{j,i}, \\
	MAE_{goaldiff} &= \frac{1}{N} \sum\limits_{i=1}^N MAE_{i}',
\end{aligned}
\]
where $N$ is the total number of matches across all tournaments. $MAE_{goals}$ is the mean absolute error for the goals, while $MAE_{goaldiff}$ is the mean absolute error between the true and predicted goal-difference.

Similar to \cite{groll:21}, we also compare our models with the bookmakers.
As the bookmakers have expert knowledge, reflected in their odds, these odds can also be used for calculating performance measures, which serve as a good benchmark compared to our models. Therefore, the  ``three-way'' odds were gathered for all matches of the UEFA EUROs 2004-2020\footnote{The  ``three-way'' odds are bookmaker odds assigned to the three ordinal match outcomes (home-team win, draw, away-team win) shortly before the start of the match. These odds were collected from the website \url{http://www.betexplorer.com/}.}, from which probabilities are calculated. For this, first $\tilde \pi_{ri}=1/\mbox{odds}_{ri}$, with $r\in\{1,2,3\}$ representing the match outcomes, is calculated for match $i$. Then, the probabilities $\hat \pi_{ri}=\tilde \pi_{ri}/c_i$ are calculated by normalizing with $c_i:=\sum_{r=1}^{3}\tilde \pi_{ri}$ to consider the bookmaker's margins, assuming that the margins are equally distributed over the three match outcomes. Having these probabilities, we can now also compute the measures ML, CR, and RPS for the bookmakers, while it is not possible to obtain measures regarding the absolute errors.

The resulting performance measures for the four models and the bookmakers are shown in Table~\ref{modelperformance}, resulting from 195 matches of the five UEFA EUROs 2004-2020\footnote{The measures are calculated for the results after 90 minutes.}. To do so, the optimal weights for the combined model had to be tuned, which is described in detail in Section~\ref{model:tuning}.
\begin{table}[ht]
	\centering
	\caption{
		\label{modelperformance} 
		Model performances regarding the multinomial likelihood, classification rate, rank probability score, and two absolute errors, with additional comparison to the bookmakers performance measures (which can only be calculated for ML, CR, and RPS). The best score of the four models is highlighted with bold font.
	}
	\begin{tabular}{lccccc}
		\specialrule{.1em}{.05em}{.05em}
		 & ML & CR & RPS & $MAE_{goals}$ & $MAE_{goaldiff}$ \\ 
		\specialrule{.1em}{.05em}{.05em}
		LASSO  & 0.3983 & 0.4872 & 0.2028 & \textbf{0.8550} & 1.1796\\
		Cforest  & \textbf{0.3996} & 0.4872 & 0.2016 & 0.8686 & \textbf{1.1669} \\
		XGBoost & 0.3738 & 0.4769 & 0.2105 & 0.8713 & 1.1968 \\ 
		Combined model  & 0.3994 & \textbf{0.4923} & \textbf{0.2015} & 0.8662 & 1.1674\\ 
		\hline
		Bookmakers  & 0.4047 & 0.5179 & 0.1973 & --- & ---\\
		\bottomrule
	\end{tabular}
\end{table}

First of all, it turns out that almost all (single) approaches achieve a slightly better performance compared to the previous analysis on
the UEFA EURO 2020 (see \citealp{groll:21}) for the ordinal performance measures (ML, CR, RPS), while for the absolute errors nearly all models could not improve. 

The Cforest slightly outperforms the other approaches with respect to the ML, closely followed by the combined model approach. On the other hand, regarding the RPS, the new combined model slightly outperforms the Cforest. Also in terms of the CR the combined model performs best. The XGBoost is significantly worse compared to all other methods and regarding all measures. However, unsurprisingly the bookmakers slightly outperform all models regarding all three performance measures. For example, the bookmakers predicted 101 out of 195 match outcomes correctly, while the combined model predicted only 96 match outcomes correctly.

In terms of the absolute errors, the Cforest and combined model yield similar results for the $MAE_{goaldiff}$, while the LASSO performs best regarding the $MAE_{goals}$.

\subsection{Tuning the weights for the combined model}\label{model:tuning}
To get the best weight-combination for the combined model, first, the three models are fitted with the best parameters obtained before.
Then, another tuning function cycles through all possible weight-combinations (with step-width of 0.05) and computes the measures ML, CR, and RPS (introduced in Section~\ref{sec:combine}) for each combination.
To compare the different combinations, each measure is then normalized and the resulting measures are averaged:
\begin{align*}
	ML_{norm} &= \frac{ML - min(ML)}{max(ML) - min(ML)} \cdot 100 \\
	CR_{norm} &= \frac{CR - min(CR)}{max(CR) - min(CR)} \cdot 100 \\
	RPS_{norm} &= \frac{max(RPS) - RPS}{max(RPS) - min(RPS)} \cdot 100 \\
	\\
	Avg_{norm} &= \frac{ML_{norm} + CR_{norm} + RPS_{norm}}{3}
\end{align*}
Using this min-max normalization, for ML and CR the largest value is normalized to 100 and for RPS the lowest value, because RPS is an error measure where a small value is best.
At the end, $Avg_{norm}$ gives on overview of the models' performance taking all measures jointly into account.

\begin{table}[ht]
	\centering
	\caption{Top-10 weight combinations and the resulting measures ordered by $Avg_{norm}$.}
	\resizebox{\textwidth}{!}{%
		\begin{tabular}{ccccccccccc}
			\toprule
			LASSO & Cforest & XGBoost & ML & CR & RPS & $ML_{norm}$ & $CR_{norm}$ & $RPS_{norm}$ & $Avg_{norm}$ \\
			\midrule
			\rowcolor{green!20} 
			0.15 & 0.85 & 0 & 0.3994 & 0.4923 & 0.2015 & 99.46 & 75 & 99.91 & 91.46 \\ 
			0.65 & 0.35 & 0 & 0.3988 & 0.4923 & 0.2019 & 97.10 & 75 & 95.35 & 89.15 \\ 
			0.7 & 0.3 & 0 & 0.3987 & 0.4923 & 0.2020 & 96.82 & 75 & 94.23 & 88.68 \\ 
			0.9 & 0.05 & 0.05 & 0.3973 & 0.4923 & 0.2027 & 91.08 & 75 & 85.87 & 83.98 \\ 
			0.35 & 0.35 & 0.3 & 0.3920 & 0.4974 & 0.2032 & 70.48 & 100 & 81.19 & 83.89 \\ 
			0.1 & 0.9 & 0 & 0.3995 & 0.4872 & 0.2015 & 99.65 & 50 & 99.71 & 83.12 \\ 
			0.2 & 0.8 & 0 & 0.3994 & 0.4872 & 0.2015 & 99.27 & 50 & 100.00 & 83.09 \\ 
			0.05 & 0.95 & 0 & 0.3995 & 0.4872 & 0.2015 & 99.83 & 50 & 99.38 & 83.07 \\ 
			0 & 1 & 0 & 0.3996 & 0.4872 & 0.2016 & 100.00 & 50 & 98.93 & 82.98 \\ 
			0.4 & 0.6 & 0 & 0.3991 & 0.4872 & 0.2015 & 98.39 & 50 & 99.14 & 82.51 \\ 
			\bottomrule
		\end{tabular}
	}
	\label{tab:weightTuning}
\end{table}
As can be seen in Table~\ref{tab:weightTuning}, the best $Avg_{norm}$ is obtained by using 15\% LASSO, 85\% Cforest, while XGBoost is completely excluded. This specific combination shows good results over all measures and, therefore, will be used for the combined model prediction. Also Table~\ref{tab:weightTuning} confirms that the Cforest model is very promising for predicting the UEFA EURO (see also \citealp{groll:21}), which is indicated by the large weight, and which one would use if the multinomial likelihood would be the most important measure.
One could also argue for other weight combinations, where the classification rate is the best (e.g.\ 35\% LASSO, 35\% Cforest, 30\% XGBoost), but this leads to worse performance regarding the multinomial likelihood.
What stands out is that the XGBoost is only included in two out of the ten best combinations at all, which could be due to the complex tuning of this method leading to worse results compared to the other methods. Moreover, our rather small sample size could also favor the LASSO and Cforest model, while the XGBoost needs more data for proper tuning, and hence, to perform as good as the other models.
As we are aiming for a good overall performance here, we decided to go with the combination leading to the best $Avg_{norm}$.

%%%%%%%%%%%%%%%%%%%%%%%%%%%%
\section{Modeling the UEFA EURO 2024}\label{sec:prediction}

In this section, we fit the combined model to our training dataset including all UEFA EUROs 2004-2020 along with the covariate data and enhanced variables. Next, we introduce a method to calculate a variable importance measure for the combined model, which gives an impression about each variable's relevance for the model. Finally, based on the eight covariates for the UEFA EURO 2024 of each participating team, the fitted combined model is used to simulate the tournament 100,000 times resulting in probabilities for each team across the tournament stages.

\subsection{Fitting the combined model to the UEFA EUROs 2004-2020 data}\label{sec:fitforest}

Next, we apply the combined model to the full dataset, which includes the five UEFA EUROs from 2004 to 2020. Therefore, each of the three models is individually fitted to this dataset.

Before fitting the models, we need to identify the optimal tuning parameters for each. For the LASSO model, the penalty parameter $\xi$ is determined using the \texttt{cv.glmnet} function from the R package \texttt{glmnet} (see \citealp{FrieEtAl:2010}). This function performs 10-fold cross-validation to find the best value for $\xi$, resulting in $\xi \approx 0.0183$.

For the Cforest model, the key parameter to tune is \texttt{mtry}, which is the size of the randomly sampled variable subset, from which one is used for splitting the node. In regression settings, it is recommended to set \texttt{mtry}$=\lfloor\sqrt{p}\rfloor=2$, with $p$ being the number of variables. As this is only a rule of thumb, alternatively, a 10-fold cross-validation can be performed over a range of \texttt{mtry} values, selecting the one that minimizes the 10-fold cross-validated error based on the negative Poisson log-likelihood. This tuning process (with \texttt{mtry}$\in \{1,2,3,4\}$) resulted in \texttt{mtry}$=1$, which we then used to fit the Cforest model to our dataset. Also the reason for choosing the tuned Cforest, resulted from the better performance when using a tuned \texttt{mtry} compared to a fixed \texttt{mtry} (see Table\ref{top8_comparison}).

Finally, for the XGBoost model, several parameters require tuning. However, as noted previously, the XGBoost model is not part of the combined model used for the final prediction, and, hence, was not derived on the full training dataset.

\subsection{Variable Importance}
\begin{figure}[h]
	\centering
	\includegraphics[width=0.7\textwidth]{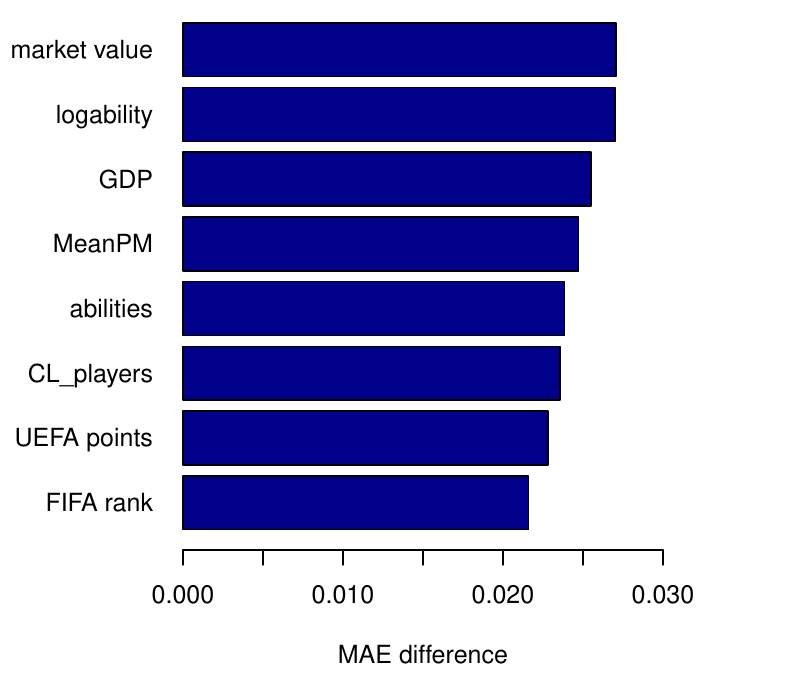}
	\captionof{figure}{$MAE_{diff}$ obtained through permutation.}
	\label{fig:mae}
\end{figure}

The variable importance is a measure, which indicates how much a feature contributes to the model for the prediction.
As the combined model is a weighted combination of three different models, there was no direct function available to obtain a variable importance for this model. Although every single model has functions for this (e.g.\ \texttt{varimp, coef, xgb.importance}), in contrast to the predictions, those can not be directly combined. This is due to the different approaches each of these functions uses to measure a variable importance.
Hence, we follow the concept of permutation-based variable importance and implemented the following approach:
\begin{enumerate}
	\item Fit the combined model on the UEFA EUROs 2004-2020 dataset (training set).
	\item Make an in-sample prediction on the training dataset.
	\item Calculate the corresponding original mean absolute error ($MAE_{org}$).
	\item For every variable used for training:
	\begin{itemize}[]
		\item Randomly permute the values of this variable. 
		\item Derive the in-sample prediction on the permuted training data.
		\item Calculate $MAE_{perm}$, as well as the difference $MAE_{diff} = MAE_{perm} - MAE_{org}$. 
	\end{itemize}
\end{enumerate}
As explained in \cite{groll:21}, permuting a variable involves randomly rearranging its values across its dataset. For instance, if we permute the feature \emph{UEFA.points}, e.g.\ the points difference between Germany and Spain in 2008 might end up assigned to the points difference of Turkey and Italy in 2020. This random reshuffling causes variables to lose their original relationships with other variables, including any associations with the response variable.
Therefore, permuting more relevant variables will result in a larger difference between the permuted error ($MAE_{perm}$) and the original error ($MAE_{org}$).
This whole procedure is repeated 100 times and the results are then averaged, to minimize the influence of randomness. %``good" permutations, which (by chance) do not destroy the information of a feature properly.
The resulting variable importance for MAE can be found in Figure~\ref{fig:mae}. It turns out that the enhanced feature logability (see Section~\ref{sec:bookmaker:data}) as well as the market value are the most important features for the model. Moreover, the GDP and the enhanced feature MeanPM have high importance for the model.

\begin{figure}[h]
	\centering
	\includegraphics[width=\textwidth]{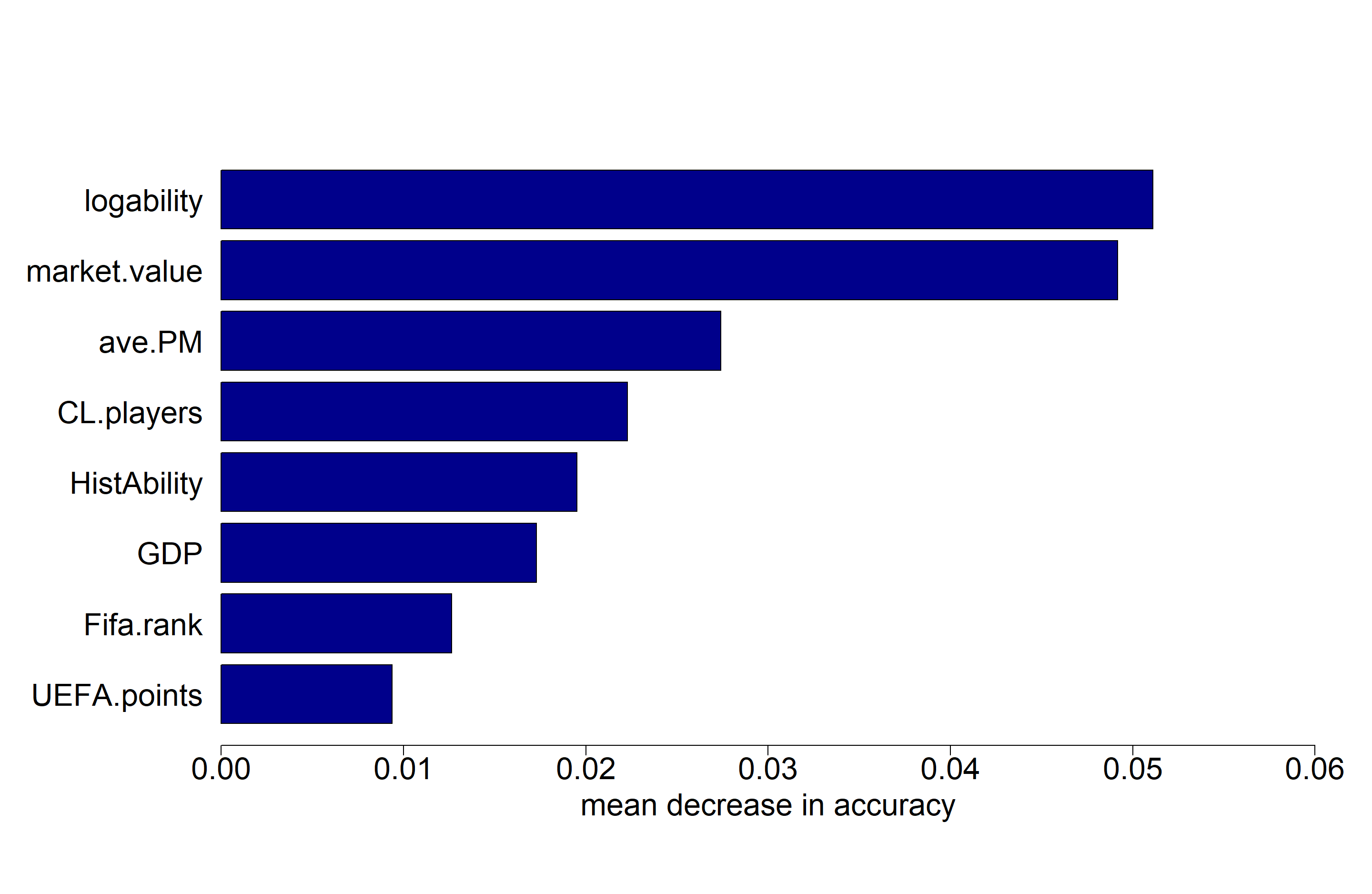}
	\captionof{figure}{Bar plot displaying the variable importance of the final fitted Cforest model used in the combined model.}
	\label{Cforest_var_imp}
\end{figure}

As the combined model mostly relies on the Cforest, it is also interesting to see the variable importance particularly for the Cforest. This is obtained using the \texttt{varimp} function \citep{Hotetal:2006}, which also uses a similar permutation-based calculation for the importance. Figure~\ref{Cforest_var_imp} shows the resulting variable importance, with the logability and market value being clearly the most important features, which is also reflected in the combined model's variable importance. Additionally, Table~\ref{LASSO_VI} shows the coefficients of the final LASSO model used in the combined model. This model surprisingly does not consider the enhanced features \textit{ave.PM} and \textit{HistAbility}, along with the \textit{UEFA.points}. For that reason, it might be useful to use a simple ridge penalty instead of LASSO. As the variables were already preselected regarding their importance, it might be unnecessary to (possibly) exclude variables by using LASSO.

Overall, these findings, along with the variable importance from Figure~\ref{Cforest_var_imp}, confirm the large impact of the enhanced variables (especially the \textit{logability}) for the models, making it worthy to estimate them in separate statistical approaches, and maybe even to introduce further enhanced variables in the future. For illustration of the differences in the resulting rankings corresponding to the three enhanced variables, together with the FIFA ranking, see Table~\ref{tab_rank}.

\begin{table}[h]
	\centering
	\captionof{table}{Final coefficients for the fitted LASSO model used in the combined model.}
	\begin{tabular}{lr||lr}
		\toprule
		\textbf{Feature} & \textbf{Coefficient} & \textbf{Feature} & \textbf{Coefficient} \\
		\midrule
		Intercept & 0.1308 & FIFA.rank & -0.0017 \\
		logability & 0.4037 & UEFA.points & 0 \\
		market.value & 0.0943 & ave.PM & 0 \\
		GDP & 0.0876 & HistAbility & 0\\
		CL.players & 0.0162 & & \\
		\bottomrule
	\end{tabular}
	\label{LASSO_VI}
\end{table}

\begin{table}[!h]
	\small
	\caption{\label{tab_rank} Comparison of the participating teams from the UEFA EURO 2024 according to rankings derived from the enhanced variables \textit{HistAbility} (Section~\ref{sec:historic}), \textit{logability} (Section~\ref{sec:bookmaker:data}), \textit{ave.PM} (Section~\ref{sec:pm:data}), and additionally the FIFA ranking.}\vspace{0.4cm}
	\centering
	\begin{tabular}{l|ll|ll|llll}
		\multicolumn{1}{c}{}&\multicolumn{2}{c}{\textbf{Historic match}} &\multicolumn{2}{c}{\textbf{Bookmaker consensus}} &\multicolumn{2}{c}{\textbf{PM player}}  &\multicolumn{2}{c}{\textbf{FIFA}}\\
		\multicolumn{1}{c}{}&\multicolumn{2}{c}{\textbf{abilities}} &\multicolumn{2}{c}{\textbf{abilities}} &\multicolumn{2}{c}{\textbf{rating}} &\multicolumn{2}{c}{\textbf{ranking}}\\
		\toprule
		\input{tab_rank}
	\end{tabular}
\end{table}
\pagebreak

\subsection{Probabilities for UEFA EURO 2024 winner} \label{sec:simul}

Now, as the combined model is fitted and trained properly, we can apply it to the current data for the UEFA EURO 2024, resulting in estimates for the expected number of goals for the competing teams, across every possible match. Therefore, shortly before the start of the UEFA EURO 2024 the covariates are collected and the enhanced abilities are estimated for each team. To simulate the tournament, we follow the same procedure as used in \cite{groll:21}.

For each match in UEFA EURO 2024, the combined model predicts the expected number of goals for both teams. Using these estimates, match outcomes are simulated based on two (conditionally) independent Poisson distributions for each team's score. This method allows us to simulate all 36 group stage matches and determine the final group standings, with respect to the official UEFA rules.

Having the final group standings, we follow the official tournament tree and simulate the knockout stage. If a knockout stage match ends with a draw after 90 minutes, we simulate 30 minutes extra time using 1/3 of the mean number of goals. If the draw persists after extra time, we randomly select the winner, reflecting a 50/50 chance for the teams to win the penalty shootout.

By repeating this simulation process of a single tournament 100,000 times, we generate probabilities for each of the 24 teams across all knockout stages, including winning probabilities. Table~\ref{winner_probs} displays these probabilities alongside with the winning probabilities derived from the various bookmakers.

\begin{table}[!h]
\small
\caption{\label{winner_probs}Resulting probabilities (in \%) for each team across all knockout-stages at the UEFA EURO 2024 estimated by simulating the the tournament 100,000 times using the final combined model. Additionally, winning probabilities based on the bookmakers odds are displayed.}\vspace{0.4cm}

\centering
\input{Winner_probs}
\end{table}

According to our combined model, the 2022 FIFA World Cup finalist France is the favored team with a predicted winning probability of 19.2\%, followed by the 2020 UEFA EURO finalist England with 16.7\%, while the host Germany has a winning probability of 13.7\%. The previous UEFA EURO 2020 champion Italy only has the seventh largest probability with 5.6\%, which to some extend also reflects the trend from the last years where Italy also failed to qualify for the World Cup 2022.

Compared to the bookmakers, there are some interesting differences between our prediction. The main difference is the fact that the bookmakers see England (20.5\%) as the favorite over France (17.6\%). Another interesting difference is that the bookmakers see Portugal slightly better than Spain compared to our prediction. Also the Netherlands have higher chances in our prediction (7.6\%), which could also be due to the very small probabilities obtained for the nations on rank 13-24. This can be seen when we aggregate the winning probabilities from the teams on rank 13-24 regarding Table~\ref{winner_probs} (3\% in our prediction vs.\ 7.4\% in the bookmakers consensus).

In addition to the probabilities of becoming European champion, Table~\ref{winner_probs} provides further interesting insights for the individual knockout stages within the tournament. For example, Austria has a significantly lower probability of reaching the Round of 16 than teams like Turkey, Ukraine or Czech Republic, but after the Round of 16 Austria has the largest probability among these teams. This is due to the rather tough group, where they are competing with France, Netherlands and Poland. Qualifying for the next round in this group is much more difficult than passing the group of Belgium, Romania, Ukraine and Slovakia (e.g.\ regarding the FIFA ranking; Table~\ref{tab_rank}). This is also shown when we compare the probability of Austria and Ukraine reaching the Round of 16 (53.4\% vs. 72.3\%) and the probability of reaching the quarterfinal (26.0\% vs. 23.3\%), which decreases much less for Austria in later stages compared to Ukraine.

It is also notable that some teams have a decent chance of reaching the round of 16, but their prospects for progressing beyond that stage decrease significantly thereafter. For example, Turkey has a promising 75.1\% chance of reaching the knockout stage, but only a moderate 23.2\% chance of progressing to the quarterfinals. Similar, teams like Denmark and Switzerland have a rather good chance of reaching the knockout stage ($\approx 75\%$), while only having a small chance of winning the tournament ($\approx 2\%$).

For Georgia and Albania it is nearly impossible to win the tournament, because they have rather weak squads (e.g.\ regarding the PM player rating; Table~\ref{tab_rank}), and Albania also was assigned to a rather tough group with Spain, Italy and Croatia.

\section{Concluding remarks}\label{sec:conclusion}

In this work, for modeling and predicting the UEFA EURO 2024, various statistical and machine learning approaches have been analyzed. These include LASSO-regularized Poisson regression, a random forest (Cforest), an extreme gradient boosting approach and, finally, the combination of these three models to one overall combined model. To train these models, the match results of the UEFA EUROs 2004-2020 were linked with covariate information of the teams as well as three highly informative enhanced variables derived from separate statistical models. These enhanced variables reflect team abilities based on historic match data of the national teams, individual performance of players or the bookmaker odds. The predictive performance of the models is evaluated using leave-one-tournament-out cross validation on the training data regarding several performance measures. The best results were obtained by the combined model (15\% LASSO + 85\% Cforest).

The combined model was then fitted to the whole training data and estimates for each possible match of the UEFA EURO 2024 were calculated. Using these estimates, the UEFA EURO 2024 was simulated 100,000 times following the official tournament tree, in order to obtain probabilities across all knockout stages. According to the simulations, France is the clear favorite with a winning probability of 19.2\%, followed by England (16.7\%) and Germany (13.7\%).

We believe there is still room for improvement in the combined model. One reason for this is that although the XGBoost model is not part of our combined model due to its weak performance, we believe this method has still some potential. This was also indicated in \cite{groll:21}, with a possible reason for the weak performance being the small size of the training data containing only four UEFA EUROs. % in the cross validation. 
Moreover, the XGBoost involves many tuning parameters, which are complex to tune. A more comprehensive tuning process could eventually lead to better results. Having a well-tuned XGBoost model could also possibly improve the combined model. %, with then all three different models included. 
Furthermore, it could be interesting to investigate the Poisson regression model using both $L_1$-regularization and $L_2$-regularization, as the variables are preselected based on their importance. Therefore, they should all have a positive impact on the model, making it (possibly) unnecessary to exclude any using LASSO. Furthermore, there is still room for improvement in the way the best weights for the combined model are obtained in using more or different performance measures. Moreover, new enhanced variables or additional informative covariates could be incorporated into the model, leading to improved results. Hence, we are looking forward to further improve these methods in the future to exploit the full potential and receive even better results.

\subsection*{Acknowledgment}
%Thanks to Andreas Groll for his exceptional support and assistance throughout the entire duration of the thesis. Also special thanks to Achim Zeileis, Lars-Magnus Hvattum, and Christophe Ley for providing the enhanced variables. Moreover, also thanks to Dominik Niedziela for the support in collecting the current covariate data for all teams and all others that supported me during the thesis. 
We thank Dominik Niedziela for his effort in helping us to collect part of the covariate data.

\pagebreak

%%%%%%%%%%%%
%%%%%%%%%%%%
\appendix
\section*{Appendix}

\section{Skellam distribution}\label{sec:general}

The Skellam distribution \citep{Ske:46} is the discrete probability distribution for a random variable $K = G_1 - G_2$, which is the difference between two independent Poisson-distributed random variables $G_1\sim Po(\lambda_1), G_2\sim Po(\lambda_2)$. Then, the probability mass function for the Skellam distribution is expressed as
$$
P(K=k)=e^{-(\lambda_1+\lambda_2)}\left(\frac{\lambda_1}{\lambda_2}\right)^{k/2}I_k(2\sqrt{\lambda_1\lambda_2}),\quad k \in \mathds{Z},
$$
where $I_k(\cdot)$ represents the modified Bessel function of the first kind. For our purposes, $G_1$ and $G_2$ represent the number of goals for two competing teams, as described in Section~\ref{sec:combine}. This allows us to calculate the three match outcome probabilities 
\begin{align}
	\text{Win team 1:} \hspace{1cm} P(G_1>G_2) &= P(K>0), \nonumber \\
	\text{Draw:} \hspace{1cm} P(G_1=G_2) &= P(K=0), \nonumber \\
	\text{Win team 2:} \hspace{1cm} P(G_1<G_2) &= P(K<0). \nonumber
\end{align}

\section{Additional bookmaker odds data}\label{sec:appendix:ata}

\begin{table}[h!]
\footnotesize
\centering
\begin{tabular}{rrrrrrrrr}
\hline
 & FRA & ITA & NED & POR & ESP & GER & ENG & CZE \\ \hline
Oddset & 3.25 & 5 & 5.5 & 6 & 6.5 & 7 & 7 & 7 \\
\hline
 & SWE & DEN & RUS & GRE & CRO & BUL & SUI & LVA \\ \hline
Oddset & 20 & 20 & 40 & 45 & 45 & 60 & 60 & 100 \\\hline
\end{tabular}
\caption{\label{tab:odds2004} Quoted odds from ODDSET for the 16~teams in the EURO~2004.}
\end{table}

\pagebreak

% latex table generated in R 4.3.3 by xtable 1.8-4 package
% Fri Jul 26 18:32:47 2024
\begin{table}[h!]
	\scriptsize
	\centering
	\begin{tabular}{rrrrrrrrrrrrr}
		\hline
		& ENG & FRA & GER & POR & ESP & ITA & NED & BEL & CRO & DEN & TUR & SUI \\ 
		\hline
		bwin & 4.33 & 4.50 & 6.00 & 8.00 & 8.50 & 17.00 & 17.00 & 19.00 & 34.00 & 41.00 & 67.00 & 67.00 \\ 
		bet365 & 4.33 & 5.00 & 6.50 & 8.00 & 9.00 & 15.00 & 17.00 & 17.00 & 41.00 & 41.00 & 51.00 & 67.00 \\ 
		Sky Bet & 4.33 & 5.00 & 6.00 & 8.50 & 9.00 & 13.00 & 19.00 & 21.00 & 34.00 & 41.00 & 67.00 & 67.00 \\ 
		Paddy Power & 4.00 & 5.00 & 5.50 & 8.50 & 8.50 & 19.00 & 19.00 & 19.00 & 41.00 & 41.00 & 67.00 & 81.00 \\ 
		William Hill & 4.50 & 5.00 & 6.50 & 8.00 & 9.00 & 19.00 & 17.00 & 21.00 & 41.00 & 41.00 & 51.00 & 81.00 \\ 
		888sport & 4.00 & 5.00 & 6.50 & 8.00 & 9.00 & 19.00 & 17.00 & 21.00 & 41.00 & 41.00 & 51.00 & 81.00 \\ 
		Betfair Sportsbook & 4.00 & 5.00 & 5.50 & 8.50 & 8.50 & 19.00 & 19.00 & 19.00 & 41.00 & 41.00 & 67.00 & 81.00 \\ 
		Bet Victor & 4.00 & 4.50 & 6.00 & 8.00 & 9.00 & 17.00 & 19.00 & 19.00 & 34.00 & 41.00 & 67.00 & 81.00 \\ 
		Coral & 4.33 & 4.50 & 5.50 & 8.00 & 8.00 & 17.00 & 17.00 & 17.00 & 34.00 & 41.00 & 67.00 & 67.00 \\ 
		Unibet & 4.25 & 5.00 & 6.50 & 8.50 & 9.00 & 17.00 & 15.00 & 21.00 & 41.00 & 41.00 & 51.00 & 67.00 \\ 
		Spreadex & 4.00 & 5.00 & 5.25 & 8.00 & 9.00 & 17.00 & 17.00 & 17.00 & 34.00 & 41.00 & 67.00 & 51.00 \\ 
		Betfred & 5.00 & 5.00 & 6.50 & 8.00 & 9.00 & 17.00 & 17.00 & 19.00 & 41.00 & 41.00 & 51.00 & 51.00 \\ 
		BetMGM UK & 4.25 & 5.00 & 6.50 & 8.50 & 9.00 & 17.00 & 15.00 & 21.00 & 41.00 & 41.00 & 51.00 & 67.00 \\ 
		Boylesports & 4.00 & 5.00 & 6.50 & 8.00 & 9.00 & 17.00 & 17.00 & 17.00 & 34.00 & 34.00 & 51.00 & 67.00 \\ 
		10Bet & 4.00 & 4.50 & 6.50 & 8.00 & 8.00 & 15.00 & 17.00 & 17.00 & 41.00 & 46.00 & 51.00 & 76.00 \\ 
		StarSports & 3.75 & 5.00 & 6.50 & 9.00 & 9.00 & 19.00 & 19.00 & 21.00 & 41.00 & 46.00 & 81.00 & 81.00 \\ 
		BetUK & 4.25 & 5.00 & 6.50 & 8.50 & 9.00 & 17.00 & 15.00 & 21.00 & 41.00 & 41.00 & 51.00 & 67.00 \\ 
		Sporting Index & 4.00 & 5.00 & 5.25 & 8.00 & 9.00 & 17.00 & 17.00 & 17.00 & 34.00 & 41.00 & 67.00 & 51.00 \\ 
		Live Score UK & 4.25 & 5.00 & 6.50 & 8.50 & 9.00 & 17.00 & 15.00 & 21.00 & 41.00 & 41.00 & 51.00 & 67.00 \\ 
		QuinnBet & 4.20 & 4.75 & 6.50 & 8.00 & 9.00 & 15.00 & 17.00 & 17.00 & 41.00 & 41.00 & 51.00 & 67.00 \\ 
		Betway & 4.33 & 5.00 & 6.50 & 8.50 & 9.00 & 17.00 & 17.00 & 17.00 & 34.00 & 41.00 & 51.00 & 81.00 \\ 
		Ladbrokes & 4.33 & 4.50 & 5.50 & 8.00 & 8.00 & 17.00 & 17.00 & 17.00 & 34.00 & 41.00 & 67.00 & 67.00 \\ 
		Midnite & 4.00 & 4.50 & 6.00 & 9.00 & 9.00 & 15.00 & 17.00 & 17.00 & 41.00 & 41.00 & 51.00 & 51.00 \\ 
		BetGoodwin & 4.20 & 4.75 & 6.50 & 8.00 & 9.00 & 15.00 & 17.00 & 17.00 & 41.00 & 41.00 & 51.00 & 67.00 \\ 
		Vbet & 4.00 & 5.00 & 6.50 & 9.00 & 9.00 & 15.00 & 15.00 & 17.00 & 41.00 & 41.00 & 51.00 & 67.00 \\ 
		AK Bets & 4.20 & 5.25 & 6.50 & 9.00 & 10.00 & 19.00 & 19.00 & 23.00 & 46.00 & 56.00 & 91.00 & 86.00 \\ 
		Betfair & 4.50 & 5.30 & 6.68 & 8.44 & 8.84 & 19.10 & 22.00 & 23.00 & 45.00 & 54.00 & 108.00 & 93.00 \\ 
		Matchbook & 4.50 & 5.30 & 6.68 & 8.44 & 8.84 & 19.10 & 22.00 & 23.00 & 45.00 & 54.00 & 108.00 & 93.00 \\ 
		\hline
		& AUT & SRB & UKR & HUN & SCO & CZE & POL & ROU & SVN & SVK & ALB & GEO \\ 
		\hline
		bwin & 51.00 & 67.00 & 101.00 & 81.00 & 101.00 & 101.00 & 101.00 & 151.00 & 201.00 & 301.00 & 301.00 & 501.00 \\ 
		bet365 & 81.00 & 81.00 & 101.00 & 81.00 & 101.00 & 151.00 & 151.00 & 201.00 & 251.00 & 501.00 & 501.00 & 501.00 \\ 
		Sky Bet & 67.00 & 67.00 & 101.00 & 101.00 & 151.00 & 81.00 & 151.00 & 201.00 & 201.00 & 251.00 & 501.00 & 501.00 \\ 
		Paddy Power & 81.00 & 81.00 & 126.00 & 126.00 & 151.00 & 201.00 & 201.00 & 201.00 & 501.00 & 501.00 & 501.00 & 501.00 \\ 
		William Hill & 81.00 & 81.00 & 101.00 & 81.00 & 101.00 & 151.00 & 101.00 & 126.00 & 251.00 & 501.00 & 301.00 & 501.00 \\ 
		888sport & 81.00 & 81.00 & 101.00 & 81.00 & 101.00 & 151.00 & 101.00 & 126.00 & 251.00 & 501.00 & 301.00 & 501.00 \\ 
		Betfair Sportsbook & 81.00 & 81.00 & 126.00 & 126.00 & 151.00 & 201.00 & 201.00 & 201.00 & 501.00 & 501.00 & 501.00 & 501.00 \\ 
		Bet Victor & 67.00 & 101.00 & 81.00 & 101.00 & 151.00 & 151.00 & 151.00 & 201.00 & 251.00 & 501.00 & 501.00 & 751.00 \\ 
		Coral & 51.00 & 67.00 & 67.00 & 81.00 & 101.00 & 101.00 & 151.00 & 151.00 & 201.00 & 251.00 & 251.00 & 501.00 \\ 
		Unibet & 81.00 & 67.00 & 101.00 & 81.00 & 101.00 & 151.00 & 151.00 & 201.00 & 251.00 & 501.00 & 501.00 & 501.00 \\ 
		Spreadex & 51.00 & 67.00 & 51.00 & 101.00 & 176.00 & 81.00 & 176.00 & 176.00 & 501.00 & 501.00 & 501.00 & 501.00 \\ 
		Betfred & 51.00 & 81.00 & 101.00 & 81.00 & 126.00 & 151.00 & 126.00 & 201.00 & 201.00 & 401.00 & 351.00 & 501.00 \\ 
		BetMGM UK & 81.00 & 67.00 & 101.00 & 81.00 & 101.00 & 151.00 & 151.00 & 201.00 & 251.00 & 501.00 & 501.00 & 501.00 \\ 
		Boylesports & 67.00 & 81.00 & 81.00 & 67.00 & 101.00 & 151.00 & 81.00 & 151.00 & 301.00 & 201.00 & 501.00 & 251.00 \\ 
		10Bet & 76.00 & 81.00 & 101.00 & 81.00 & 101.00 & 151.00 & 151.00 & 201.00 & 251.00 & 451.00 & 451.00 & 501.00 \\ 
		StarSports & 81.00 & 81.00 & 126.00 & 151.00 & 151.00 & 176.00 & 176.00 & 301.00 & 501.00 & 501.00 & 751.00 & 501.00 \\ 
		BetUK & 81.00 & 67.00 & 101.00 & 81.00 & 101.00 & 151.00 & 151.00 & 201.00 & 251.00 & 501.00 & 501.00 & 501.00 \\ 
		Sporting Index & 51.00 & 67.00 & 51.00 & 101.00 & 176.00 & 81.00 & 176.00 & 176.00 & 501.00 & 501.00 & 501.00 & 501.00 \\ 
		Live Score UK & 81.00 & 67.00 & 101.00 & 81.00 & 101.00 & 151.00 & 151.00 & 201.00 & 251.00 & 501.00 & 501.00 & 501.00 \\ 
		QuinnBet & 81.00 & 81.00 & 101.00 & 81.00 & 101.00 & 151.00 & 151.00 & 201.00 & 251.00 & 451.00 & 451.00 & 501.00 \\ 
		Betway & 81.00 & 81.00 & 101.00 & 81.00 & 101.00 & 151.00 & 151.00 & 201.00 & 251.00 & 201.00 & 351.00 & 501.00 \\ 
		Ladbrokes & 51.00 & 67.00 & 67.00 & 81.00 & 101.00 & 101.00 & 151.00 & 151.00 & 201.00 & 251.00 & 251.00 & 501.00 \\ 
		Midnite & 76.00 & 71.00 & 101.00 & 81.00 & 71.00 & 101.00 & 101.00 & 201.00 & 251.00 & 501.00 & 251.00 & 501.00 \\ 
		BetGoodwin & 81.00 & 81.00 & 101.00 & 81.00 & 101.00 & 151.00 & 151.00 & 201.00 & 251.00 & 451.00 & 451.00 & 501.00 \\ 
		Vbet & 81.00 & 67.00 & 101.00 & 81.00 & 81.00 & 151.00 & 151.00 & 201.00 & 251.00 & 501.00 & 501.00 & 501.00 \\ 
		AK Bets & 86.00 & 111.00 & 111.00 & 151.00 & 176.00 & 176.00 & 201.00 & 176.00 & 626.00 & 626.00 & 751.00 & 751.00 \\ 
		Betfair & 83.00 & 127.00 & 127.00 & 147.00 & 245.00 & 265.00 & 255.00 & 196.00 & 843.00 & 941.00 & 902.00 & 911.00 \\ 
		Matchbook & 83.00 & 127.00 & 127.00 & 147.00 & 245.00 & 265.00 & 255.00 & 186.00 & 843.00 & 951.00 & 921.00 & 911.00 \\
	\end{tabular}
	\caption{\label{tab:odds2020} Quoted winning odds from 28~online bookmakers
		for the 24~teams in the EURO~2024 obtained on 2024-06-09 from
		\url{https://www.oddschecker.com/} and \url{https://www.bwin.com/}, respectively.}
\end{table}

\pagebreak

\bibliographystyle{apalike}
\bibliography{literatur}

\end{document}

%% file: tab_rank.tex
% latex table generated in R 4.0.3 by xtable 1.8-4 package
% Mon Jun  7 17:12:26 2021
 1 & \includegraphics[width=0.4cm]{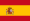} & Spain & \includegraphics[width=0.4cm]{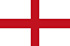} & England & \includegraphics[width=0.4cm]{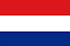} & Netherlands & \includegraphics[width=0.4cm]{FRA.png} & France \\ 
  2 & \includegraphics[width=0.4cm]{POR.png} & Portugal & \includegraphics[width=0.4cm]{FRA.png} & France & \includegraphics[width=0.4cm]{FRA.png} & France & \includegraphics[width=0.4cm]{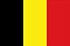} & Belgium \\ 
  3 & \includegraphics[width=0.4cm]{FRA.png} & France & \includegraphics[width=0.4cm]{GER.png} & Germany & \includegraphics[width=0.4cm]{GER.png} & Germany & \includegraphics[width=0.4cm]{ENG.png} & England \\ 
  4 & \includegraphics[width=0.4cm]{ENG.png} & England & \includegraphics[width=0.4cm]{POR.png} & Portugal & \includegraphics[width=0.4cm]{ENG.png} & England & \includegraphics[width=0.4cm]{POR.png} & Portugal \\ 
  5 & \includegraphics[width=0.4cm]{BEL.png} & Belgium & \includegraphics[width=0.4cm]{ESP.png} & Spain & \includegraphics[width=0.4cm]{POR.png} & Portugal & \includegraphics[width=0.4cm]{NED.png} & Netherlands \\ 
  6 & \includegraphics[width=0.4cm]{NED.png} & Netherlands & \includegraphics[width=0.4cm]{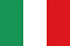} & Italy & \includegraphics[width=0.4cm]{ESP.png} & Spain & \includegraphics[width=0.4cm]{ESP.png} & Spain \\ 
  7 & \includegraphics[width=0.4cm]{GER.png} & Germany & \includegraphics[width=0.4cm]{NED.png} & Netherlands & \includegraphics[width=0.4cm]{BEL.png} & Belgium & \includegraphics[width=0.4cm]{ITA.png} & Italy \\ 
  8 & \includegraphics[width=0.4cm]{ITA.png} & Italy & \includegraphics[width=0.4cm]{BEL.png} & Belgium & \includegraphics[width=0.4cm]{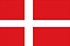} & Denmark & \includegraphics[width=0.4cm]{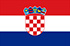} & Croatia \\ 
  9 & \includegraphics[width=0.4cm]{CRO.png} & Croatia & \includegraphics[width=0.4cm]{CRO.png} & Croatia & \includegraphics[width=0.4cm]{ITA.png} & Italy & \includegraphics[width=0.4cm]{GER.png} & Germany \\ 
  10 & \includegraphics[width=0.4cm]{DEN.png} & Denmark & \includegraphics[width=0.4cm]{DEN.png} & Denmark & \includegraphics[width=0.4cm]{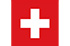} & Switzerland & \includegraphics[width=0.4cm]{SUI.png} & Switzerland \\ 
  11 & \includegraphics[width=0.4cm]{SUI.png} & Switzerland & \includegraphics[width=0.4cm]{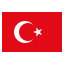} & Turkey & \includegraphics[width=0.4cm]{TUR.png} & Turkey & \includegraphics[width=0.4cm]{DEN.png} & Denmark \\ 
  12 & \includegraphics[width=0.4cm]{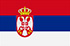} & Serbia & \includegraphics[width=0.4cm]{SUI.png} & Switzerland & \includegraphics[width=0.4cm]{CRO.png} & Croatia & \includegraphics[width=0.4cm]{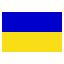} & Ukraine \\ 
  13 & \includegraphics[width=0.4cm]{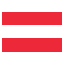} & Austria & \includegraphics[width=0.4cm]{AUT.png} & Austria & \includegraphics[width=0.4cm]{SRB.png} & Serbia & \includegraphics[width=0.4cm]{AUT.png} & Austria \\ 
  14 & \includegraphics[width=0.4cm]{UKR.png} & Ukraine & \includegraphics[width=0.4cm]{SRB.png} & Serbia & \includegraphics[width=0.4cm]{UKR.png} & Ukraine & \includegraphics[width=0.4cm]{HUN.png} & Hungary \\ 
  15 & \includegraphics[width=0.4cm]{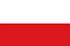} & Poland & \includegraphics[width=0.4cm]{HUN.png} & Hungary & \includegraphics[width=0.4cm]{AUT.png} & Austria & \includegraphics[width=0.4cm]{POL.png} & Poland \\ 
  16 & \includegraphics[width=0.4cm]{HUN.png} & Hungary & \includegraphics[width=0.4cm]{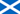} & Scotland & \includegraphics[width=0.4cm]{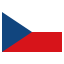} & Czech Republic & \includegraphics[width=0.4cm]{SRB.png} & Serbia \\ 
  17 & \includegraphics[width=0.4cm]{CZE.png} & Czech Republic & \includegraphics[width=0.4cm]{UKR.png} & Ukraine & \includegraphics[width=0.4cm]{POL.png} & Poland & \includegraphics[width=0.4cm]{CZE.png} & Czech Republic \\ 
  18 & \includegraphics[width=0.4cm]{TUR.png} & Turkey & \includegraphics[width=0.4cm]{CZE.png} & Czech Republic & \includegraphics[width=0.4cm]{SCO.png} & Scotland & \includegraphics[width=0.4cm]{SCO.png} & Scotland \\ 
  19 & \includegraphics[width=0.4cm]{SCO.png} & Scotland & \includegraphics[width=0.4cm]{POL.png} & Poland & \includegraphics[width=0.4cm]{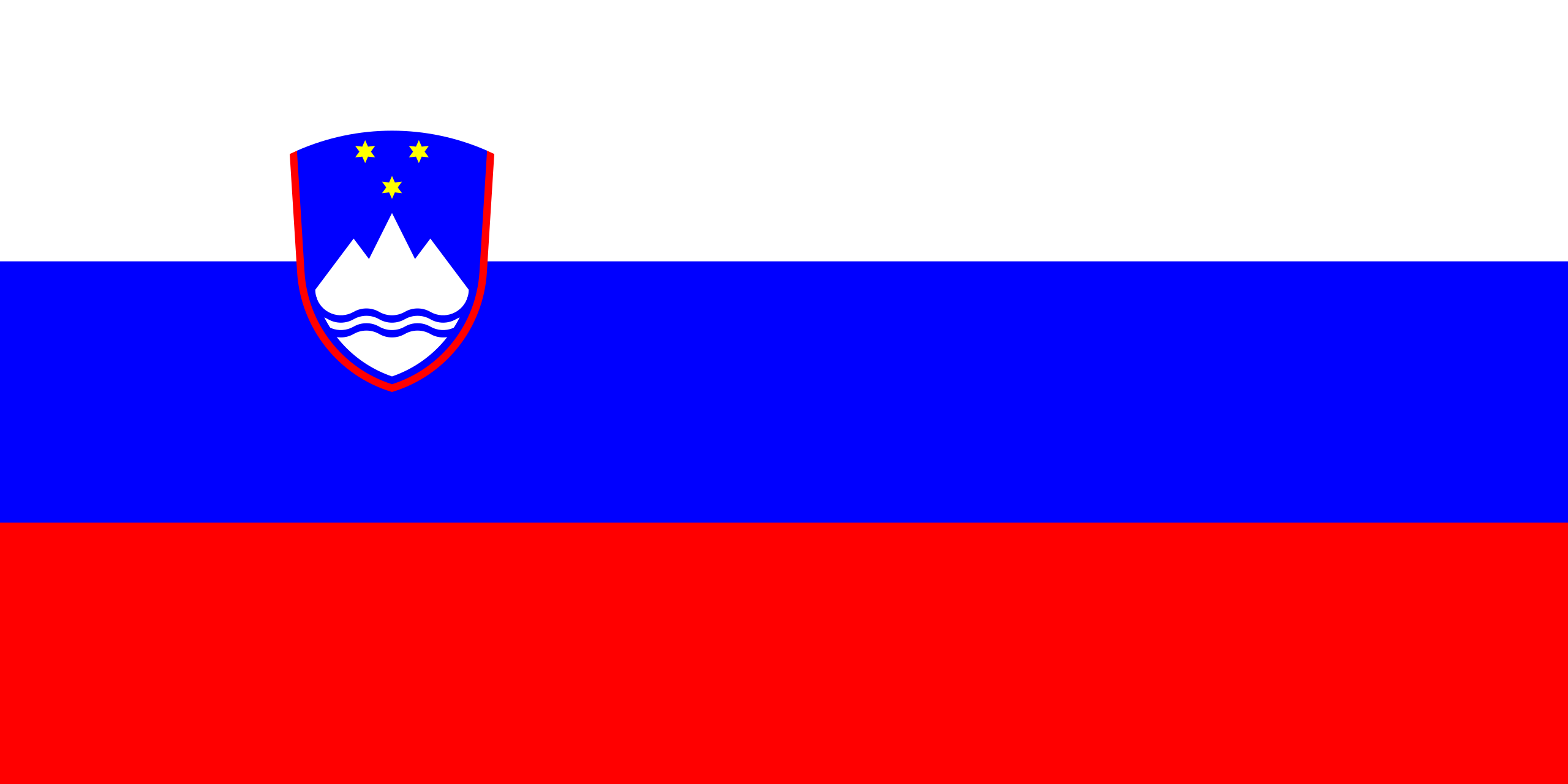} & Slovenia & \includegraphics[width=0.4cm]{TUR.png} & Turkey \\ 
  20 & \includegraphics[width=0.4cm]{SVN.png} & Slovenia & \includegraphics[width=0.4cm]{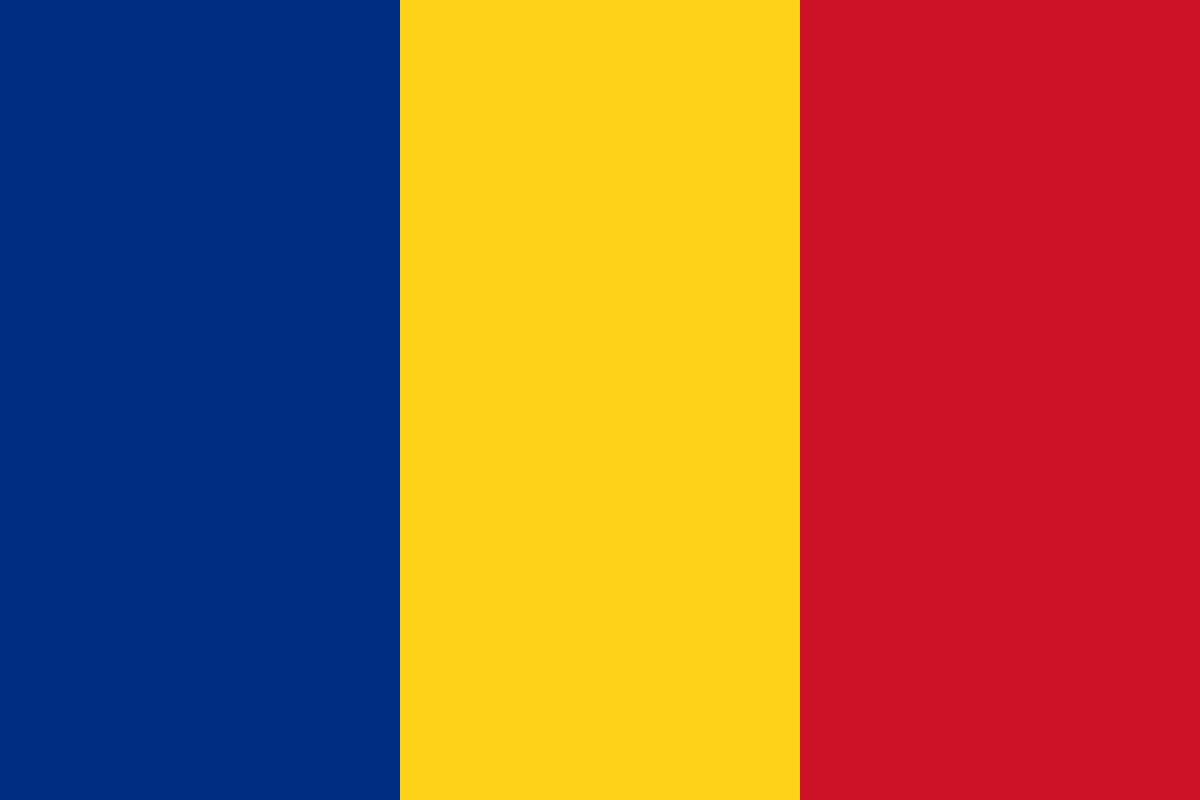} & Romania & \includegraphics[width=0.4cm]{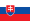} & Slovakia & \includegraphics[width=0.4cm]{ROU.png} & Romania \\ 
  21 & \includegraphics[width=0.4cm]{ROU.png} & Romania & \includegraphics[width=0.4cm]{SVN.png} & Slovenia & \includegraphics[width=0.4cm]{HUN.png} & Hungary & \includegraphics[width=0.4cm]{SVK.png} & Slovakia \\ 
  22 & \includegraphics[width=0.4cm]{SVK.png} & Slovakia & \includegraphics[width=0.4cm]{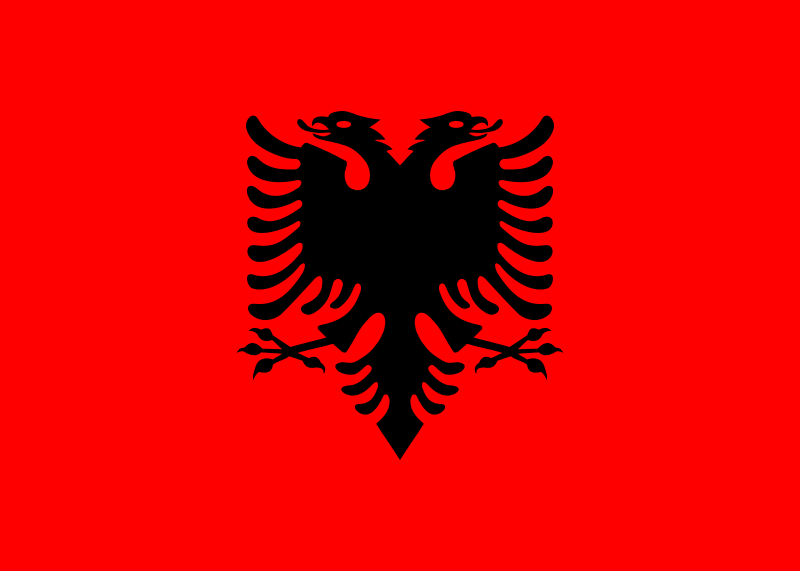} & Albania & \includegraphics[width=0.4cm]{ROU.png} & Romania & \includegraphics[width=0.4cm]{SVN.png} & Slovenia \\ 
  23 & \includegraphics[width=0.4cm]{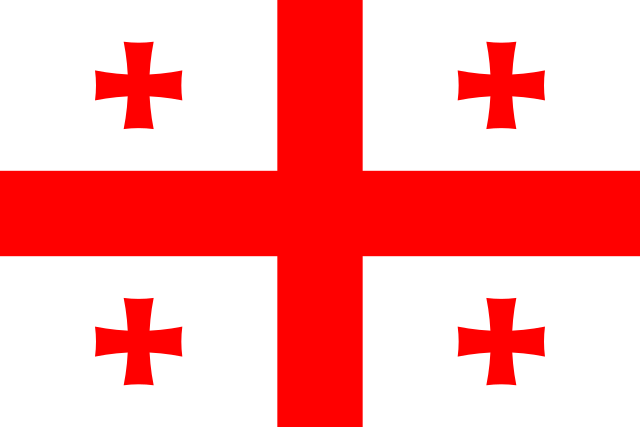} & Georgia & \includegraphics[width=0.4cm]{SVK.png} & Slovakia & \includegraphics[width=0.4cm]{ALB.png} & Albania & \includegraphics[width=0.4cm]{ALB.png} & Albania \\ 
  24 & \includegraphics[width=0.4cm]{ALB.png} & Albania & \includegraphics[width=0.4cm]{GEO.png} & Georgia & \includegraphics[width=0.4cm]{GEO.png} & Georgia & \includegraphics[width=0.4cm]{GEO.png} & Georgia \\ 
  

%% file: Winner_probs.tex
% latex table generated in R 4.3.3 by xtable 1.8-4 package
% Sun Jun  9 17:21:33 2024
\begin{tabular}{lllrrrrrr}
   \toprule
 &  & & Round & Quarter & Semi & Final & European & Bookmakers \\ 
   &  & & of 16 & finals & finals &  & Champion & consensus \\ 
   \midrule
1. & \includegraphics[width=0.4cm]{FRA.png} & FRA & 93.6 & 72.6 & 50.2 & 31.2 & 19.2 & 17.6 \\ 
  2. & \includegraphics[width=0.4cm]{ENG.png} & ENG & 97.0 & 72.2 & 47.8 & 28.8 & 16.7 & 20.5 \\ 
  3. & \includegraphics[width=0.4cm]{GER.png} & GER & 95.2 & 65.7 & 40.8 & 24.6 & 13.7 & 14.0 \\ 
  4. & \includegraphics[width=0.4cm]{ESP.png} & ESP & 93.5 & 66.6 & 38.4 & 21.6 & 11.4 & 9.7 \\ 
  5. & \includegraphics[width=0.4cm]{POR.png} & POR & 95.9 & 60.4 & 37.1 & 20.4 & 10.8 & 10.1 \\ 
  6. & \includegraphics[width=0.4cm]{NED.png} & NED & 82.1 & 55.3 & 31.6 & 16.1 & 7.6 & 4.8 \\ 
  7. & \includegraphics[width=0.4cm]{ITA.png} & ITA & 86.8 & 54.0 & 26.1 & 12.6 & 5.6 & 4.9 \\ 
  8. & \includegraphics[width=0.4cm]{BEL.png} & BEL & 93.8 & 51.6 & 24.5 & 11.2 & 4.9 & 4.5 \\ 
  9. & \includegraphics[width=0.4cm]{CRO.png} & CRO & 71.9 & 37.6 & 15.0 & 5.9 & 2.1 & 2.2 \\ 
  10. & \includegraphics[width=0.4cm]{DEN.png} & DEN & 75.5 & 35.2 & 15.0 & 6.0 & 2.0 & 2.0 \\ 
  11. & \includegraphics[width=0.4cm]{SUI.png} & SUI & 75.3 & 33.4 & 12.9 & 4.9 & 1.6 & 1.2 \\ 
  12. & \includegraphics[width=0.4cm]{AUT.png} & AUT & 53.4 & 26.0 & 10.6 & 3.6 & 1.1 & 1.1 \\ 
  13. & \includegraphics[width=0.4cm]{TUR.png} & TUR & 75.1 & 23.2 & 8.6 & 2.6 & 0.7 & 1.5 \\ 
  14. & \includegraphics[width=0.4cm]{UKR.png} & UKR & 72.3 & 23.3 & 7.1 & 1.9 & 0.4 & 0.9 \\ 
  15. & \includegraphics[width=0.4cm]{SRB.png} & SRB & 54.4 & 18.3 & 5.9 & 1.7 & 0.4 & 1.0 \\ 
  16. & \includegraphics[width=0.4cm]{CZE.png} & CZE & 66.0 & 17.8 & 6.1 & 1.6 & 0.4 & 0.6 \\ 
  17. & \includegraphics[width=0.4cm]{SCO.png} & SCO & 48.4 & 16.1 & 4.6 & 1.3 & 0.3 & 0.7 \\ 
  18. & \includegraphics[width=0.4cm]{POL.png} & POL & 38.2 & 14.9 & 4.6 & 1.2 & 0.3 & 0.5 \\ 
  19. & \includegraphics[width=0.4cm]{HUN.png} & HUN & 46.8 & 14.9 & 4.2 & 1.1 & 0.2 & 0.8 \\ 
  20. & \includegraphics[width=0.4cm]{SVN.png} & SVN & 35.0 & 9.0 & 2.1 & 0.4 & 0.1 & 0.3 \\ 
  21. & \includegraphics[width=0.4cm]{SVK.png} & SVK & 53.0 & 12.3 & 2.7 & 0.5 & 0.1 & 0.2 \\ 
  22. & \includegraphics[width=0.4cm]{ROU.png} & ROU & 48.4 & 11.1 & 2.4 & 0.5 & 0.1 & 0.5 \\ 
  23. & \includegraphics[width=0.4cm]{GEO.png} & GEO & 30.0 & 4.4 & 0.9 & 0.1 & 0.0 & 0.2 \\ 
  24. & \includegraphics[width=0.4cm]{ALB.png} & ALB & 18.3 & 4.2 & 0.7 & 0.1 & 0.0 & 0.2 \\ 
   \bottomrule
\end{tabular}